\documentclass[sigconf,nonacm]{acmart}

\usepackage{graphicx}
\usepackage{amsmath}
\usepackage{booktabs}
\usepackage{epigraph}
\usepackage{algorithm}
\usepackage{algpseudocode}
\usepackage{comment}
\usepackage{enumitem}
\setlist{nolistsep}
\usepackage{dblfloatfix}
\usepackage{subcaption}
\usepackage{hyperref}
\hypersetup{breaklinks,colorlinks}

\usepackage[capitalize]{cleveref}
\crefname{section}{Sec.}{Secs.}
\crefname{section}{Section}{Sections}
\crefname{table}{Table}{Tables}
\crefname{table}{Tab.}{Tabs.}

\newcommand{\RNum}[1]{\uppercase\expandafter{\romannumeral #1\relax}}

\newcommand{\specialcell}[2][l]{%
  \begin{tabular}[#1]{@{}l@{}}#2\end{tabular}}

\newcommand{\tabitem}{~~\llap{\textbullet}~~}

\AtBeginDocument{%
  \providecommand\BibTeX{{%
    \normalfont B\kern-0.5em{\scshape i\kern-0.25em b}\kern-0.8em\TeX}}}

\begin{document}

\title{FaceDiffuser: Speech-Driven 3D Facial Animation Synthesis Using Diffusion}

\author{Stefan Stan}
\affiliation{%
  \institution{Utrecht University}
  \city{Utrecht}
  \country{The Netherlands}}
\email{st.stan96@gmail.com}

\author{Kazi Injamamul Haque}
\affiliation{%
  \institution{Utrecht University}
  \city{Utrecht}
  \country{The Netherlands}}
\email{k.i.haque@uu.nl}

\author{Zerrin Yumak}
\affiliation{%
  \institution{Utrecht University}
  \city{Utrecht}
  \country{The Netherlands}}
\email{z.yumak@uu.nl}


\begin{abstract}
Speech-driven 3D facial animation synthesis has been a challenging task both in industry and research. Recent methods mostly focus on deterministic deep learning methods meaning that given a speech input, the output is always the same. However, in reality, the non-verbal facial cues that reside
throughout the face are non-deterministic in nature. In addition, majority of the approaches focus on 3D vertex based datasets and methods that are compatible with existing facial animation pipelines with rigged characters is scarce. To eliminate these issues, we present FaceDiffuser, a non-deterministic deep learning model to generate speech-driven facial animations that is trained with both 3D vertex and blendshape based datasets. Our method is based on the diffusion technique and uses the pre-trained large speech representation model HuBERT to encode the audio input. To the best of our knowledge, we are the first to employ the diffusion method for the task of speech-driven 3D facial animation synthesis. We have run extensive objective and subjective analyses and show that our approach achieves better or comparable results in comparison to the state-of-the-art methods.  We also introduce a new in-house dataset that is based on a blendshape based rigged character. We recommend watching the accompanying supplementary video. The code and the dataset will be released publicly\footnote[1]{\url{https://github.com/uuembodiedsocialai/FaceDiffuser}}.
\end{abstract}



\begin{teaserfigure}
\centering
  \includegraphics[width=0.89\textwidth]{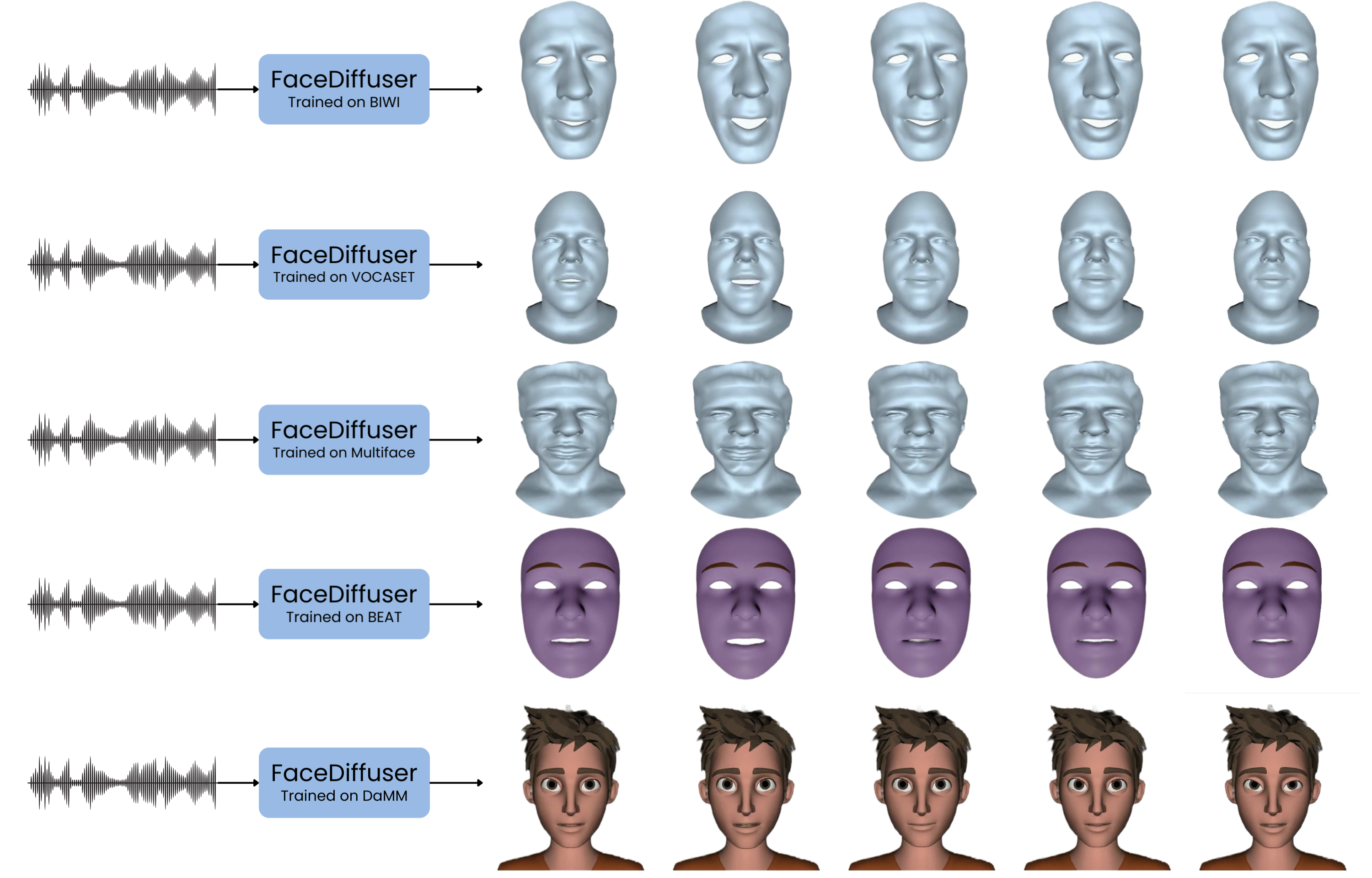}
  \caption{We present FaceDiffuser, an end-to-end non-deterministic neural network architecture for speech-driven 3D facial animation synthesis. Our proposed approach produces realistic and diverse animation sequences and is generalizable to both temporal 3D vertex based mesh animation datasets (top 3 rows) and temporal blendshape based datasets (bottom 2 rows).}
  \Description{FaceDiffuser takes audio as input and synthesises speech-driven 3D facial animations in terms of both temporal 3D vertex data and blendshape data or Maya facial control data. The generated animation sequences are expressive and diverse.}
  \label{fig:teaser}
\end{teaserfigure}

\maketitle

\section{Introduction}
3D facial animation is an important component in various applications such as gaming and XR. Generating facial animations is a tedious task, often done by experienced technical animators using keyframes or done by capturing and retargeting a performer's expression to a rigged model with blendshapes and facial controls. The former requires time and expertise to achieve while the latter requires specialised capture systems \cite{Dynamixyz, Faceware, DI4D}. Recent pipelines such as Unreal Engine MetaHuman Animator \cite{MHA} aims to provide more scalable facial animation capture solutions. Research on speech-driven 3D facial animation can be divided into phoneme-based procedural \cite{JALI, Charalambous2019} and data-driven approaches including recent deep learning based methods using intermediary representations of speech units \cite{taylor2017deep,zhou2018visemenet} and end-to-end deep learning \cite{karras2017audio, cudeiro2019voca, richard2021meshtalk,fan2022faceformer, xing2023codetalker, facexhubert2023} eliminating the need for intermediary representations. Phoneme-based methods require explicit definition of co-articulation rules and manual work. Deep learning methods automatically discover patterns that can map new speech input to output animations. Our work is an end-to-end speech-driven 3D animation method that uses a diffusion based deep learning model to generate facial animations non-deterministically.   

End-to-end deep learning based approaches effectively generate whole face animation producing promising results for accurate lip-sync and upper face animations. Karras et al. \cite{karras2017audio} proposes an end-to-end CNN based method mapping input waveforms to the 3D vertex coordinates. VOCA \cite{cudeiro2019voca} proposes a CNN based approach that takes advantage of the pre-trained DeepSpeech \cite{hannun2014deepspeech} model including identity control. MeshTalk \cite{richard2021meshtalk} learns a latent space representation of facial expressions by employing a U-Net autoencoder network and generates animations based on audio together with a template mesh as input. With the success of transformers, Fan et al. \cite{fan2022faceformer} proposed FaceFormer, a self-supervised representation learning model which employs Wav2Vec 2.0 as a speech encoder. Haque and Yumak \cite{facexhubert2023} proposed FaceXHuBERT, an efficient end-to-end encoder-decoder model based on the large speech model HuBERT \cite{hubert} including identity and emotion control. None of these methods take into account the non-deterministic nature of facial animations meaning given the same speech input, they produce the same results. However, the non-verbal facial cues that reside throughout the face are non-deterministic in nature \cite{ng2022learning2listen}. 

There are a few works that employ non-deterministic methods. Learning2Listen \cite{ng2022learning2listen} generates facial animation for the listening party in two-party dyadic interactions and uses a Vector Quantised Variational Auto Encoder (VQ-VAE) \cite{vq-vae}. CodeTalker \cite{xing2023codetalker} incorporates self-supervised Wav2Vec 2.0 inspired by FaceFormer \cite{fan2022faceformer} together with the idea of having a latent codebook using VQ-VAE inspired by \cite{ng2022learning2listen}. TalkShow \cite{yi2023generating} also employs VQ-VAE to generate both face, upper body and hand animations. Recent work on non-deterministic body motion synthesis such as \cite{tevet2022human, tseng2023edge, alexanderson2023listen} employs diffusion models. However, diffusion models have not yet been used in the speech-driven 3D facial animation domain and to our knowledge we are the first to do that. Our model employs a HuBERT \cite{hubert} audio encoder together with a specialised diffusion model and produces facial animation both for 3D vertex based and rigged characters. \cref{fig:teaser} shows animations generated with our model using five different datasets. Our extensive objective and subjective analysis shows that our model produces better or comparable results in comparison to state-of-the-art methods. 
The main contributions of our work are enumated below: 
\begin{itemize}[leftmargin=*]
    \item FaceDiffuser is the first to incorporate the diffusion mechanism for speech-driven 3D facial animation synthesis task.
    \item Our model performs better than the state-of-the-art methods in terms of objective metrics on multiple temporal high dimensional 3D vertex based mesh animation datasets. 
    \item We extend our approach to show that the proposed model can generalise to lower dimensional blendshape and facial control based datasets. A new in-house built facial controls based facial animation dataset for rigged characters is also introduced. 
    \item Extensive qualitative analysis and ablation studies were presented to demonstrate the importance of the diffusion mechanism and the ability to synthesise high-quality, diverse facial animation sequences with a discussion on the capabilities and limitations of deterministic and non-deterministic approaches.
\end{itemize}

\section{Related Work}

Speech-driven facial animation can be classified into two categories: (i) neural rendering of 2D talking heads which resides in the pixel space \cite{Lu2021, Xinya2022eamm, zhang2023sadtalker} and (ii) 3D speech-driven animation synthesis using temporal 3D vertex data \cite{cudeiro2019voca, richard2021meshtalk, fan2022faceformer, xing2023codetalker, facexhubert2023} and blendshape data for a rigged character \cite{Voice2FaceEA, zhou2018visemenet}. Another line of research focuses on 3D reconstruction of faces from 2D videos \cite{RingNet:CVPR:2019, DECA:Siggraph2021, EMOCA:CVPR:2021} however they are not speech-driven. For an extensive survey on 3D face reconstruction, tracking and morphable models, we refer to \cite{MORALES2021100400, Egger2020}.

In this paper, we focus on the 3D speech-driven facial animation synthesis problem using a diffusion model. Therefore in the following two sub-sections, we first present the state-of-the-art on speech-driven 3D facial animation and then motion synthesis using diffusion models.

\subsection{Speech-driven 3D Facial Animation}
3D speech-driven facial animation typically uses phoneme-based procedural approaches \cite{JALI, Charalambous2019}. Although these methods come with the advantage of animation control and easy integration to artist-friendly pipelines, they are not fully automatic and require defining explicit rules for co-articulation. Another line of research uses machine learning \cite{Taylor2012} or graph-based approaches \cite{Cao2005} to learn speech-animation mappings from data. These methods rely on blending between speech units and cannot capture the complexity of the dynamics of visual speech \cite{taylor2017deep}. Recent approaches on 3D speech animation synthesis effectively employ deep learning models \cite{taylor2017deep, karras2017audio, richard2021meshtalk, zhou2018visemenet, cudeiro2019voca, fan2022faceformer, facexhubert2023, Voice2FaceEA}. Taylor et al. \cite{taylor2017deep} proposes a sliding window approach instead of an RNN focusing on capturing coarticulation effects. VisemeNet \cite{zhou2018visemenet} builds upon the viseme-based JALI \cite{JALI} model and combines this with an LSTM-based neural network. However, these two methods \cite{taylor2017deep, zhou2018visemenet} still rely on intermediary representations of phonemes and they focus on the mouth movement only. Most previous works do not include automatic tongue animation except \cite{Voice2FaceEA}. Some methods use 3D face reconstruction methods from in-the-wild videos to generate their data, e.g. dyadic speech-driven facial animation \cite{jonell2020letsfaceit, ng2022learning2listen}. However, these methods are prone to 3D reconstruction errors. Most of the deep learning based approaches are based on 3D vertex based datasets \cite{karras2017audio, richard2021meshtalk, cudeiro2019voca, fan2022faceformer, facexhubert2023} and are not compatible with traditional animation pipelines with rigged characters except a few examples such as \cite{taylor2017deep, zhou2018visemenet, Voice2FaceEA}. 

Closest to our work are \cite{taylor2017deep, karras2017audio, richard2021meshtalk, zhou2018visemenet, cudeiro2019voca, fan2022faceformer, facexhubert2023, Voice2FaceEA}. Karras et al. \cite{karras2017audio} proposed an end-to-end method using CNNs aiming to resolve the ambiguity in mapping between audio and face by introducing an additional emotion component to the network. However, the method is not trained on multiple speakers and cannot handle identity variations. Instead, Cudeiro et al. \cite{cudeiro2019voca} presents the audio-driven facial animation method VOCA that generalizes to new speakers eliminating the need for retargeting. However, VOCA fails to realistically synthesise upper face motion and does not include emotional variations. Richard et al. \cite{richard2021meshtalk} aims for audio-driven animation that can capture variations in multiple speakers including a large dataset of subjects. They address the problem of lack of upper face motions using a categorical latent space that disentangles audio-correlated and audio-uncorrelated information based on a cross-modality loss. Fan et al. \cite{Fan2022audiotext} proposes an audio and text-driven facial animation method that incorporates the large language model GPT-2 \cite{gpt2}. FaceFormer \cite{fan2022faceformer} uses a self-supervised pretrained speech model that addresses the scarcity of available data in existing audio-visual datasets. The model uses a modified version of transformers to handle longer sequences of data. FaceXHuBERT \cite{facexhubert2023} proposes a more efficient network and incorporates HuBERT \cite{hubert} as the audio encoder as well as includes identity and emotion control.

However, these methods do not take into account the non- deterministic nature of facial animations. Learning2Listen \cite{ng2022learning2listen} generates facial expressions non-deterministically in two-party dyadic interactions and uses a Vector Quantised Variational Auto Encoder (VQ-VAE) \cite{vq-vae} to generate facial animation for the listening party. CodeTalker \cite{xing2023codetalker} incorporates self-supervised Wav2Vec 2.0 inspired by FaceFormer \cite{fan2022faceformer} and a modified version of VQ-VAE inspired by Learning2Listen \cite{ng2022learning2listen}. TalkShow \cite{yi2023generating} also employs VQ-VAE to generate face, upper body and hand animations. None of these methods incorporate the diffusion process to generate a variety of 3D facial animations driven by speech input.

\begin{figure*}
  \centering
  \includegraphics[width=0.98\linewidth]{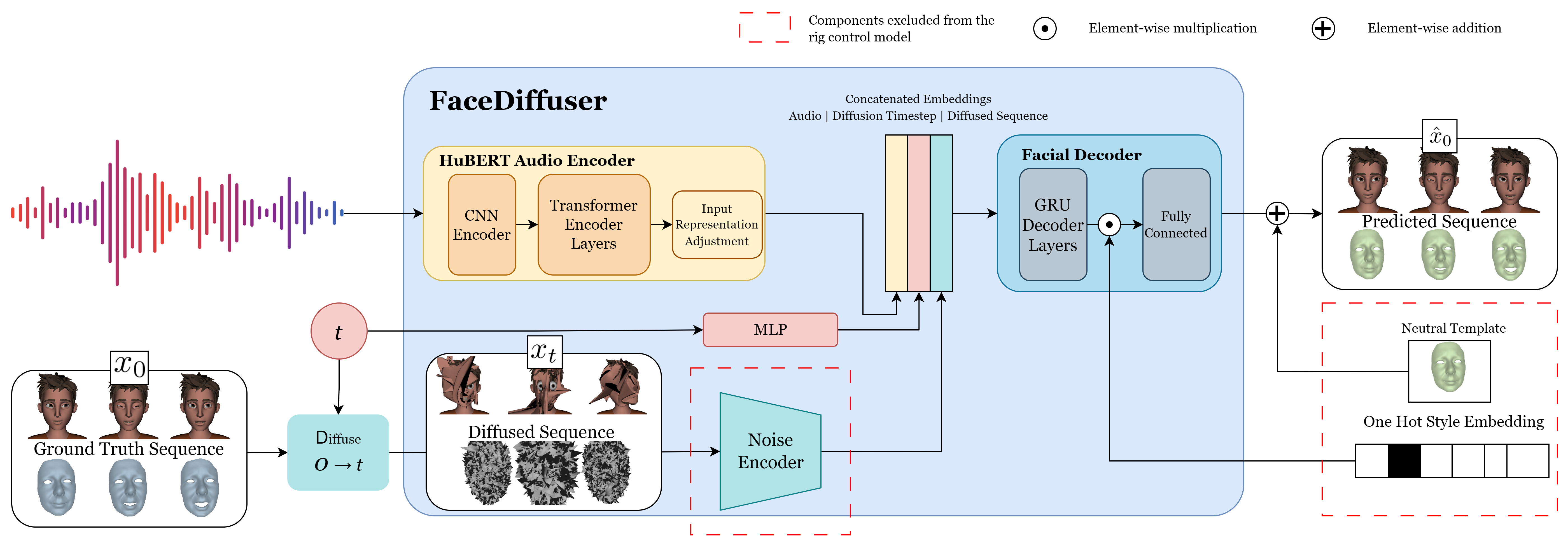}
  \caption{FaceDiffuser learns to denoise facial expressions and generate animations based on input speech. Audio speech embeddings from pre-trained HuBERT model combined with embeddings from the noised ground truth animation sequence are used to train the Facial Decoder. The Facial Decoder is comprised of a sequence of GRU layers followed by a fully connected layer and learns to predict (i) vertex displacements or (ii) rig control (blendshape) values. The predicted sequence $\hat{x}_0$ is compared with the ground truth sequence $x_0$ by computing the loss, which is then backpropagated to update the model parameters.}
  \Description{FaceDiffuser detailed figure.}
  \label{fig:ProposedApproach}
\end{figure*}

\subsection{Diffusion for Motion Synthesis}
The concept of diffusion was introduced in 2015 by Sohl-Dickstein et al. \cite{sohl2015diffusion} and is based on a concept in non-equilibrium thermodynamics. The idea is that a sample from the data distribution is gradually noised by the diffusion process and then a neural
model learns the reverse process of gradually denoising the sample \cite{tevet2022human}. It is widely used in the computer vision domain by denoising images noised e.g. with a Gaussian noise and a neural network is trained to reverse the diffusion process \cite{HoDiffusion2020, song2021scorebased}. It was used successfully in text-to-image generation tasks leading to examples such as DALL-E2 \cite{dalle2} and StableDiffusion \cite{Rombach_2022_CVPR}. For a survey of diffusion models applied in the image domain, we refer to \cite{croitoru2023diffusion}.

3D body motion generation and 2D talking face generation are the closest work we found in the literature to our work with respect to the use of diffusion process. In the domain of 3D body motion synthesis, Tevet et al. \cite{tevet2022human} proposed Human Motion Diffusion Model (MDM), a model that can generate body animations based on text descriptions. They employ a transformer-based architecture and introduce the noised ground truth motions as an additional input to the network. By doing this, they succeed in generating non-deterministic animations at inference time. With the success of MDM, other works generate body animations given music and audio as input \cite{tseng2023edge, yang2023diffusestylegesture, alexanderson2023listen}. Tseng et al. \cite{tseng2023edge} train a model that can generate dance animations conditioned on music, while Yang et al. \cite{yang2023diffusestylegesture} use a similar approach for speech-driven gesture motion synthesis. Alexanderson et al. \cite{alexanderson2023listen} apply diffusion both for co-speech gesture and dance motion generation. In the domain of 2D talking faces, a speech driven video editing method is proposed by Bigioi et al. \cite{bigioi2023speech}. By taking a template video as input along with a new speech segment, the model generates new lip motions that follows the target speech sequence. The model is capable of generalising across different speaker identities. DAE-Talker \cite{du2023dae} makes use of diffusion for generating talking head animation with a 2-stage learning process. They employ a diffusion autoencoder approach initially introduced by Preechakul et al. \cite{preechakul2022diffae} on images and extend it to generating videos. They first train a diffusion autoencoder that learns the latent space of facial expressions from the training data. The first stage has no temporal awareness and only learns to reconstruct an image from its encoding and its noised representation. In the second stage, a transformer-based encoder-decoder architecture is used to encode the audio input and output frame embeddings. To our knowledge, no work in the literature apply diffusion models for the 3D speech-driven facial animation task.

\section{Problem Formulation}
Let $A$ be an audio input associated with a sequence of ground truth frames $x_0^{1:N}=(x_0^1, x_0^2, ..., x_0^N)$, where $N$ is the number of visual frames sampled at a certain FPS specific to datasets. Each frame in the sequence $x_0^n$ is represented as an array of vertex positions with the length $V$ x $3$, where $V$ is the number of mesh vertices in the topology, and $3$ is the number of spatial axes. In the case of blendshape or facial control datasets, $x_0^n$ represents a vector of rig controls or blendshape values, having the length $C$, the number of controls or blendshapes driving the facial rig.
Based on audio input $A$, the goal of our architecture is to predict an animation sequence, $\hat{x}_0^{1:N}$ that resemble the ground truth frames $x_0^{1:N}$. Additionally, the predictions will be guided by a style $S$ in the form of a one-hot vector with a length equal to the number of training subjects (for vertex based datasets only in our experiments) and noise $x_T^{1:N}$ sampled from the normal distribution $\mathcal{N}(0, 1)$ and with the same shape as the ground truth sequence $x_0^{1:N}$. 
It is to be noted that, models trained on vertex data generates animations in the form of vertex displacements with respect to neutral face templates. Whereas, models trained on blendshape or facial control does not require this step as the neutral face of the rigged face is not a variable in our setting. 

The abstraction of the formulated problem is presented with the following equation:
\begin{equation}
    \hat{x}_0 = \text{FaceDiffuser}(\mathcal{A}, x_t, t)
\end{equation}

Where, $\hat{x}_0$ is the predicted animation sequence, $\mathcal{A}$ is the input audio sequence and $x_t$ is the sequence $x_0$ after t diffusion steps. Here, $t = T, x_t = \sigma^{1:N}$ drawn from $\mathcal{N}(0, 1)$.

\section{Proposed Approach}
\label{sec:model_arch}
\subsection{Training}
We propose a general model that can be employed for both vertex based and blendshape based datasets with slight modifications in terms of hyperparameters. We refer to the vertex based model configuration as V-FaceDiffuser, and to the blendshape based model as B-FaceDiffuser. The main difference being the additional \textit{Noise Encoder} as it can be seen in \cref{fig:ProposedApproach} enclosed in dashed red box. The noise encoder helps to project high dimensional vertex data into low dimensional latent representation. The diffusion noising process takes in $x_0^{1:N}$ to compute noised $x_t^{1:N}$, retaining its original shape. 

From \cref{fig:ProposedApproach}, we can identify the following main components that are included in both versions of the model:

\textbf{Audio Encoder:} We use the pretrained large speech model, HuBERT as the audio encoder similar to \cite{facexhubert2023} and it is kept the same in both versions of the architecture. We employ a pre-trained version of the HuBERT architecture and use the released \textit{hubert-base-ls960} version of it, which was trained on 960 hours of LibriSpeech\cite{librispeech} dataset.

\textbf{Diffusion Process:} Let $x_0^{1:N}$ be a sequence of ground truth visual frames from the dataset with shape $(N, C)$, where $C$ is either the number of vertices, multiplied by the 3 (for 3 spatial axes), or the number of rig facial controls (or blendshape values). During training, we randomly sample an integer timestep $t$ from $[1, T]$, indicating the number of noising steps to be applied to $x_0^{1:N}$ and to obtain $x_t^{1:N}$ with the formula:
\begin{equation}
    x_t^{1:N} = q(x_t^{1:N}|x_{t-1}^{1:N}) = \mathcal{N}(\sqrt{1 - \beta_t}x_{t-1}^{1:N}, (\beta_t) \mathcal{I})
\end{equation}

Where, $N$ is the number of visual frames in the sequence, $t$ is the diffusion timestep and $\beta_t$ is the constant noise at timestep $t$ such that $0 < \beta_1 < \beta_2 < ... < \beta_T < 1$.  

After the forward noising process, ideally, we want to be able to compute the reverse process and go backwards from $x_T^{1:N} \sim \mathcal{N}(0, 1)$ to $x_0^{1:N}$. Therefore the conditional distribution function $p(x_{t-1}^{1:N}|x_{t}^{1:N})$ needs to be known beforehand. \textit{Ho et al.} \cite{ho2020denoising} proposes to achieve that by learning the latent representation variance of the dataset. The training objective is defined as learning to predict the noise $\epsilon$ that was added to the input $x_0$. 
However, we deviate from \cite{ho2020denoising} and follow MDM \cite{tevet2022human} and EDGE \cite{tseng2023edge}, choosing for our model to learn to predict actual animation data instead of the noise level in the data. We consider this to be more suitable for our task since the results are also conditioned on the input audio. Furthermore, by choosing this approach, our model is able to predict acceptable results even from the first denoising steps of the inference process, allowing for faster sampling. However following the full inference process would give the best results.

We employ a simple loss for training similar to \cite{tevet2022human} and \cite{tseng2023edge}. More thorough experimentation was conducted by \textit{Ho et al.} \cite{ho2020denoising}, who also claim that utilising the simple loss for learning the variational bound proved to be both easier to implement and also advantageous for the quality of the sampled results. The loss is defined as:

\begin{equation}
    \mathcal{L} = E_{x_0 \sim q(x_0|c), t \sim [1, T]}[\|x_0 - \hat{x}_0 \|]
\end{equation}

\textbf{Facial Decoder:} The facial decoder is responsible for producing the final animation frames from latent representation of the encoded audio and noise. It is comprised of multiple GRU layers followed by a final fully connected layer that predicts the output sequence. During the decoding step, a style embedding can also be added in the form of an element-wise product between a learned style embedding vector and the hidden states output. We explain the choice of the GRU decoder in the ablation section.

\subsection{Inference}
\begin{figure}[t]
  \centering
   \includegraphics[width=0.95\linewidth]{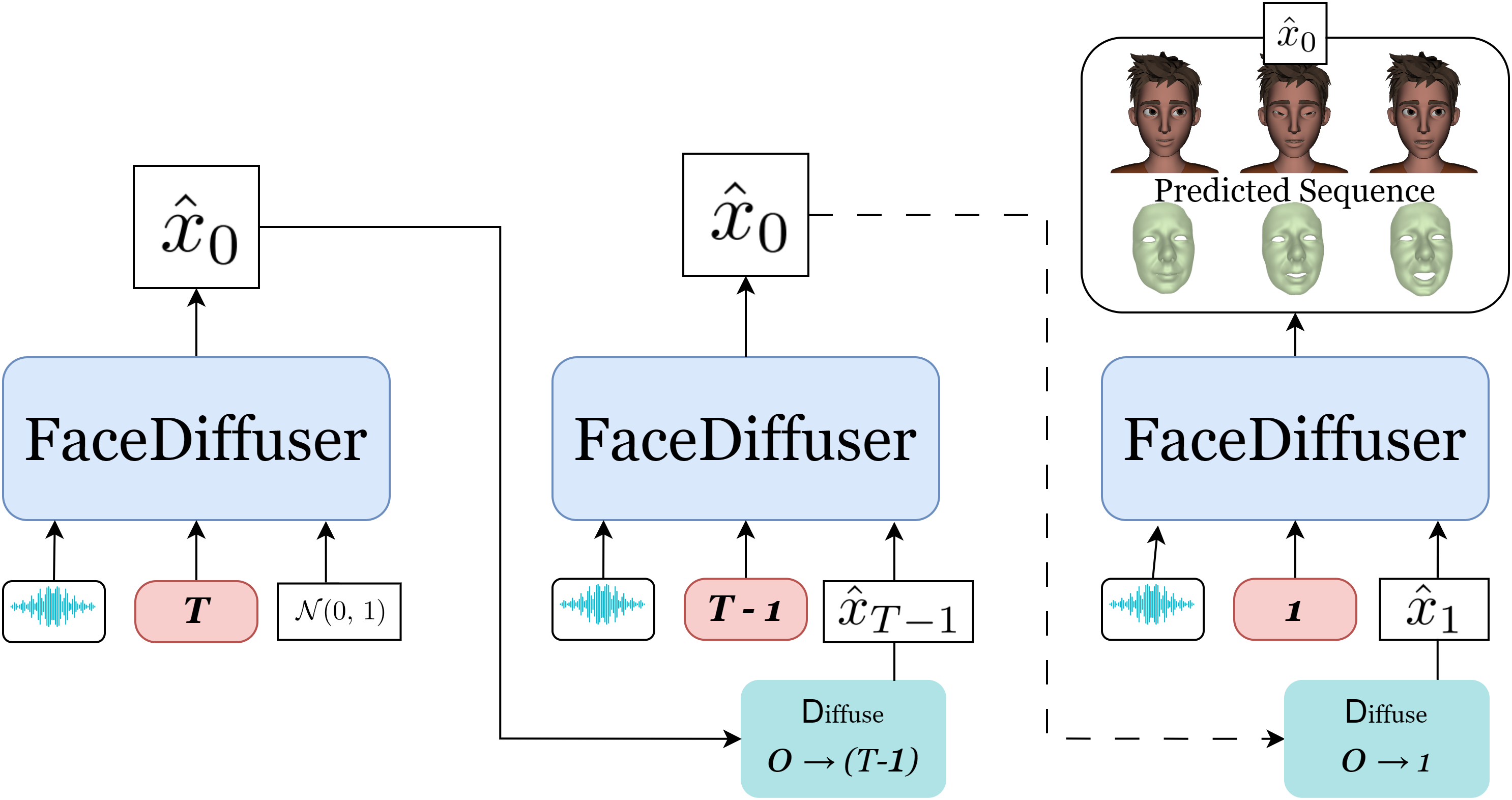}
   \caption{FaceDiffuser inference is an iterative process from T decreasing to 1. The initial noise being represented by actual noise from the normal distribution $\mathcal{N}(0, 1)$. At each step, we provide the network with the audio and noised animation input. The predicted motion is then diffused again and fed to the next step of the iteration.}
   \Description{}
   \label{fig:model_inference}
\end{figure}

The inference is an iterative process, during which we go through all of the diffusion timesteps backwards from $T$ to $1$, gradually improving the prediction at each inference step. During inference, the ground truth noised sequence as input to the model is missing compared to training time. Therefore, at inference time $T$, a randomly sampled noise is provided to the model from the uniform distribution $\mathcal{N}(0, 1)$. \cref{fig:model_inference} depicts the inference process.

\section{Experimental Setup}
For our experiments, we trained our model on 3D vertex datasets. We use BIWI \cite{fanelli20103biwi} as our primary vertex based dataset for comparisons against the state-of-the-art methods such as VOCA \cite{cudeiro2019voca}, FaceFormer \cite{fan2022faceformer} and CodeTalker \cite{xing2023codetalker}. The dataset contains audio-4D scan pairs of 14 subjects, each uttering 40 sentences twice- with neutral emotion and with emotional expressions. We adopt the exact same dataset split for BIWI as done in \cite{fan2022faceformer, xing2023codetalker} and only use the emotional sequences subset. The split results into training set \textit{BIWI-Train} that contains 192 sentences and validation set \textit{BIWI-Val} that contains 24 sentences from 6 training subjects. There are two test sets: BIWI-Test-A, containing 24 sentences from seen subjects and BIWI-Test-B, containing 32 sentences from the 8 remaining unseen subjects. The former test set is used for computing objective metrics while both facilitate the qualitative analysis and the perception study. BIWI-Test-B is further used to compute the diversity metric that we define in the next section. We also train all the methods on VOCASET, adopting the VOCASET-Train and VOCASET-Test split following \cite{cudeiro2019voca, fan2022faceformer, xing2023codetalker} and similar to these works, we use the test set to generate animations for the perceptual user study. In addition, we also employ the Multiface dataset \cite{wuu2022multiface} which was also used in MeshTalk \cite{richard2021meshtalk} to demonstrate the generalisability capability of our proposed method and compare our results to two state-of-the-art methods. Among these three datasets, only Multiface has proper eye-blinks while for VOCASET, there are no examples of eye-blinks and for BIWI, the face topology does not contain proper eye-lids. In our experiments, we used BIWI and MultiFace for objective analysis and BIWI and VOCASET for perceptual studies with users. For the qualitative analysis, all datasets were used. For clarity, we defined the models trained on vertex data in V-FaceDiffuser configuration.

\begin{table}[t]
    \centering
    \begin{tabular}{ccc}
         \toprule
         Hyperparameter &  V-FaceDiffuser & B-FaceDiffuser \\
         \midrule
         Optimizer & Adam & Adam \\
         Learning Rate & $1e^{-4}$ & $1e^{-4}$ \\
         Number of epochs & 50 & 100 \\
         Diffusion Steps &  500 & 1000 \\
         $\beta$ schedule & linear & linear \\
         Input Embedding Dim & 512(B); 256(V,M) & N/A \\
         Number of GRU Layers & 2 & 4(U);2(B) \\
         GRU hidden size & 512 & 1024(U); 256(B)\\
         GRU dropout & 0.3 & 0.3 \\
         \bottomrule
    \end{tabular}
    \caption{Hyperparameter values of our proposed approach. For V-FaceDiffuser, since the vertex data is high dimensional, we embed it to a latent dimension. The input embedding dimension is 512 for BIWI (represented as 512(B)) while for VOCASET and Multiface, it is 256 (i.e.- 256(V,M)). B-FaceDiffuser does not need this projection as the data is low dimensional. Different number of GRU layers and hidden sizes were used for BEAT and UUDaMM, denoted by (B) and (U) respectively.}
    \label{tab:hyperparam_tuning}
\end{table}

In addition, we trained our proposed method on blendshape based datasets such as BEAT \cite{liu2021beat} and our in-house dataset, UUDaMM (Utrecht University Dyadic Multimodal Motion Capture Dataset) for a rigged character. While both datasets include both face and body animations, we use the facial data together with the synchonously captured audio data only. While BEAT facial animations are based on Apple ARKIT 52 blendshape standard, UUDaMM dataset includes AutoDesk Maya facial controls. These are much lower dimensional datasets compared to the vertex based ones. Since there has not yet been any speech-driven facial animation work done with these two datasets in the literature to our knowledge, we compare the results of our approach with a baseline method that is identical to the proposed method without the diffusion component. Moreover, because the compared state-of-the-art methods are designed and proposed for vertex based datasets specifically, employing those on the blendshape datasets for direct comparison is not applicable. For BEAT, we use a subset ($\approx$16 hour data by native English speakers) of the full ($\approx$76 hours) dataset. UUDaMM ($\approx$ 10 hours) consists of multimodal motion capture data of 2 actors interacting in a natural dyadic setting. The dataset contains full body motion, captured with Vicon \cite{Vicon}, facial performance capture with Dynamixyz \cite{Dynamixyz}, synchronised audio recording and reference videos. The facial performance data was solved and retargeted to a publicly available model - Ray\cite{RayCharacterMaya}. We then export the temporal facial control values (where the facial controls drive the artist generated blendshapes) to form the training dataset. We defined the models trained on lower dimensional blendshape based datasets as B-FaceDiffuser. Eye-blinks are present in both these datasets while UUDaMM also includes eye gaze in the training data. More details on these 5 datasets we used are available in the supplementary material.

\textbf{Implementation Details:} All the model training was done on a shared compute cluster running Linux with AMD EPYC 7313 CPU, Nvidia A16 GPU, 1TB RAM. \cref{tab:hyperparam_tuning} shows the hyperparameters we use for the proposed approaches. 

\section{Results}
We evaluate our proposed approach quantitatively, qualitatively and with a perceptual user study and compared our results to the state-of-the-art methods. In the following subsections, we will present and discuss the results. 

\subsection{Quantitative Evaluation}
\label{sec:quant_eval}
Following \cite{fan2022faceformer, xing2023codetalker, facexhubert2023}, we employ the lip vertex error (LVE for V-FaceDiffuser, LBE for B-FaceDiffuser) in order to measure the lip-sync error. Similar to \cite{xing2023codetalker}, we also adopt the FDD metric that gives an indication of the upper face motion variation in terms of statistics and how close it is to the variation observed in the ground truth. Additionally, we also compute mean full-face vertex error (MVE for V-FaceDiffuser, MBE for B-FaceDiffuser) as we are interested in not only the lip-sync but also the motion that resides throughout the entire face. We use the exact same set of lip and upper face vertices to compute the mentioned metrics as provided in code repositories of \cite{xing2023codetalker} and \cite{richard2021meshtalk} for BIWI and Multiface respectively. In order to demonstrate the diversity capability of our proposed model, we introduce a novel diversity metric that is subject to animation generated conditioned on different training subjects for the vertex based datasets. For the blendshape based datasets, we manually select the blendshapes related to lip and upper face movements for LBE and FDD respectively. 

\begin{table}[t]
    \centering
    \resizebox{\linewidth}{!}{
    \begin{tabular}{cccccc}
    \toprule
        Dataset & Method & MVE $\downarrow$ & LVE $\downarrow$ & FDD $\downarrow$  & Diversity$\uparrow$\\
         &  &x$10^{-3}$ & x$10^{-4}$  & x$10^{-5}$& x$10^{-3}$\\
         &  & mm &  mm  & mm & mm\\
    \midrule
         & VOCA & 8.3606 & 6.7155 & 7.5320 & 7.8507\\
         & FaceFormer & 7.1658 & 4.9847 & 5.0972 & 5.9201\\
         & CodeTalker & 7.3980 & 4.7914 & 4.1170 & 0.0003\\
         BIWI & V-FaceDiffuser & \textbf{6.8088} & \textbf{4.2985} & \textbf{3.9101} & \textbf{9.2459} \\
         \midrule
        & VOCA & 15.782 & 25.067 & 14.253 & 0.5292\\
         & FaceFormer & 7.6132 & 7.0770 & \textbf{5.0127} & 14.745\\
         & CodeTalker & 12.170 & 20.392 & 6.6857 & 13.423\\
         Multiface & V-FaceDiffuser & \textbf{7.0004} & \textbf{6.2295} & 5.9020 & \textbf{15.500} \\
         
    \bottomrule
    \end{tabular}
    } 
    \caption{Objective results computed over the temporal 3D vertex datasets. Our approach achieves the best results in all four objective metrics for the BIWI dataset. For the Multiface dataset we score best on all metrics except the FDD metric. }
    \label{tab:obj_vertex_data}
\end{table}

\paragraph{\textbf{MVE and MBE}} Mean vertex (or blendshape) error measures the deviation of all the face vertices (or all the blendshapes) of a sequence with respect to the ground truth by computing the maximal L2 error for each frame and by taking a mean over all corresponding frame pairs. 

\paragraph{\textbf{LVE and LBE}} Lip vertex (or blendshape) error is identical to MVE (or MBE). We only consider the lip vertices (or blendshapes related to lips) for computing the metrics.  

\paragraph{\textbf{FDD}} Introduced in \cite{xing2023codetalker}, it measures the variation of facial dynamics for a motion sequence against ground truth. It gives an indication of how close the standard deviation (or upper face motion variation) of a generated sequence is compared to the observed variation in ground truth.

\begin{table}[t]
    \centering
    \begin{tabular}{ccccc}
    \toprule
        Dataset & Method & MBE $\downarrow$ & LBE $\downarrow$ & FDD $\downarrow$ \\
    \midrule
         & w/o Diffusion & \textbf{0.4170} & \textbf{0.1077} & 0.1482 \\
         BEAT & B-FaceDiffuser & 0.5152 & 0.1358 & \textbf{0.1471}  \\
         \midrule
         & w/o Diffusion & \textbf{2.6963} & \textbf{1.5924} & 2.0553 \\
         UUDaMM & B-FaceDiffuser & 3.4479 & 1.6671 & \textbf{1.7752}\\
    \bottomrule
    \end{tabular}
    \caption{Objective evaluation results computed over the blendshape and controller based datasets. Best results are marked as \textbf{bold}. Our model generates slightly higher error based on frame-level low dimensional GT values while achieving better FDD.}
    \label{tab:obj_blendshape_data}
\end{table}

\paragraph{\textbf{Diversity}} We introduce an additional metric that measures the model's ability to produce diverse animations. With the introduction of diffusion, our goal is to develop a model able to generate a great range of motions and non-deterministic expressions for the regions of the face that are uncorrelated or loosely correlated to speech. Based on similar metrics used for diffusion models in literature \cite{tevet2022human}, we introduce a novel diversity metric for the face. 
The diversity metric could be applied to different iterations of the inference algorithm, however, to be compatible with deterministic state-of-the-art baselines, we define the diversity across different subjects. Therefore, we define the diversity metric as follows. Let $\hat{x}_0^s$ be a generated sequence, conditioned on subject $s \in S$, where $S$ is the list of training subjects. We compute the mean vertex difference between every 2 sequences conditioned on different subjects. We then take the mean of these differences and define the diversity of a sequence as follows:
\begin{equation}
    Diversity = \frac{\sum_{i=1}^{|S| - 1}\sum_{j=i+1}^{|S|} \|\hat{x}_0^i - \hat{x}_0^j \|}{\frac{(|S| - 1) \cdot |S| }{2}}
\end{equation}
where $S$ denotes the list of the training subjects and $\hat{x}_0^i$ is the predicted animation sequence conditioned on the $i$th subject from $S$. 
In our training setting, the diversity metric is only suitable for the vertex based datasets that allow using neutral face template meshes that are different across subjects. This ensures different subjects have different neutral facial physiognomy. Whereas for the blendshape based datasets, the neutral configuration of the blendshape values remains identical for different actors/speakers. Hence, B-FaceDiffuser is trained without the subject conditioning and we evaluate the diversity in terms of animation graphs.

\begin{figure}[t]
  \centering
   \includegraphics[width=0.95\linewidth]{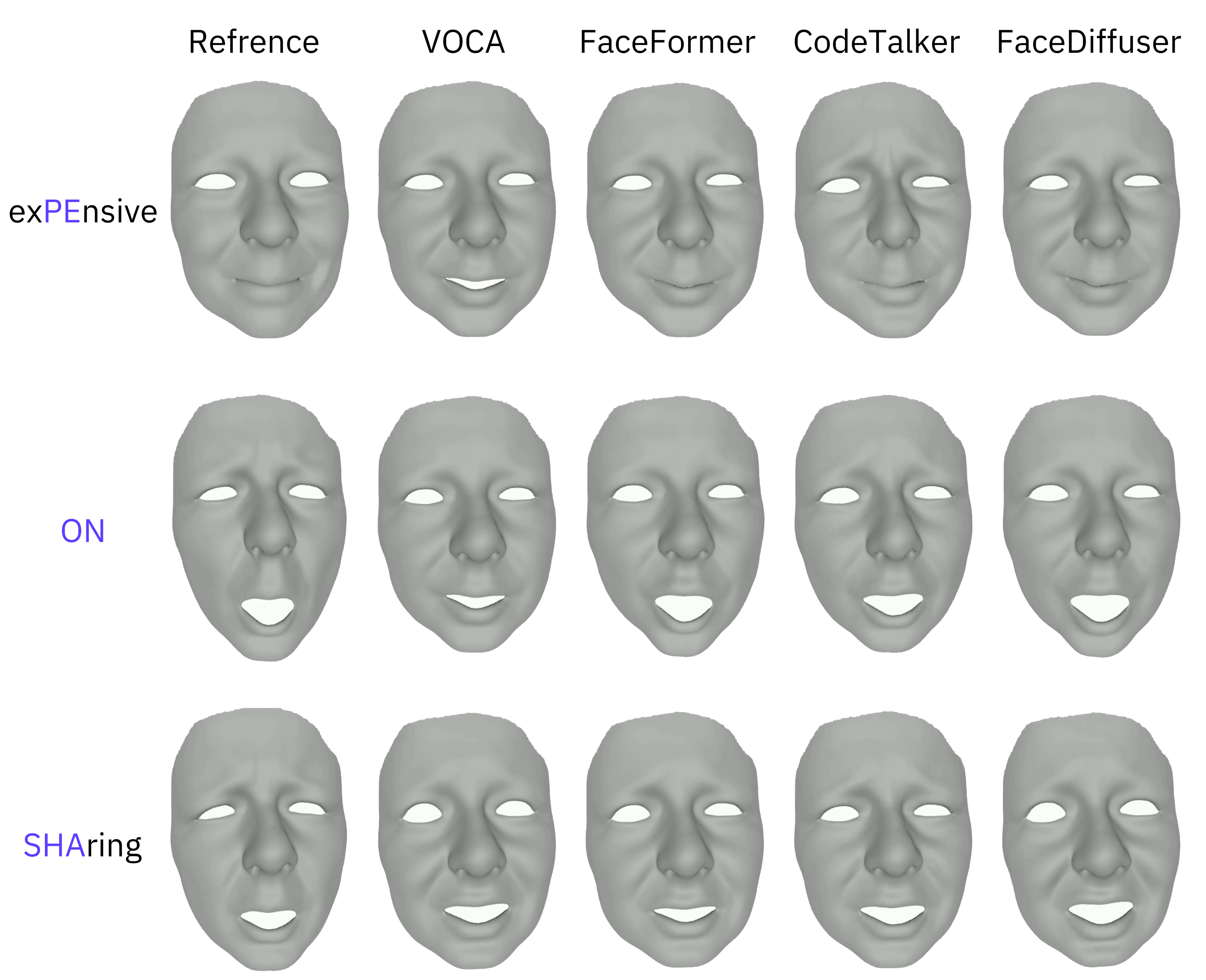}
   \caption{Visual comparison of frames from synthesised facial animation sequences generated by different methods together with reference frames from GT sequence. The highlighted utterances are represented in visual frames. Our method generates lip shapes that are close to the references while encouraging diversity in the upper face.}
   \Description{}
   \label{fig:biwi_speech_comp}
   \vspace{-5mm}
\end{figure}

\paragraph{\textbf{Discussion}} \cref{tab:obj_vertex_data} and \cref{tab:obj_blendshape_data} show the objective results for V-FaceDiffuser and B-FaceDiffuser respectively. Our approach performs better than all the other methods on all the objective metrics for the BIWI dataset. For the Multiface dataset, ours perform the best on all objective metrics but FDD, for which FaceFormer performs slightly better. For the blendshape based datasets, we cannot compare our method with state-of-the-art methods as mentioned earlier. Instead, we compare the diffusion  MBE and LBE are higher than the baseline model as our approach encourages randomness and non-determinism while more resembling the upper face variation observed in the ground truth with lower FDD value. Furthermore, the diversity for B-FaceDiffuser is evaluated qualitatively in \cref{sec:qualitative}.

\subsection{Qualitative Evaluation}
\label{sec:qualitative}
We carried out extensive qualitative analysis of the generated animation sequences and compared them to both ground truth and other methods. Our method generates accurate lip shapes resembling ground truth lip motions while generating diverse upper-face motions. A visual comparison can be observed in \cref{fig:biwi_speech_comp}. Moreover, our approach is generalisable to unseen speakers, noisy audio input, multiple overlapping speakers in audio, speech in different languages and text-to-speech audio.   

\begin{figure}[t]
  \centering
   \includegraphics[width=0.95\linewidth]{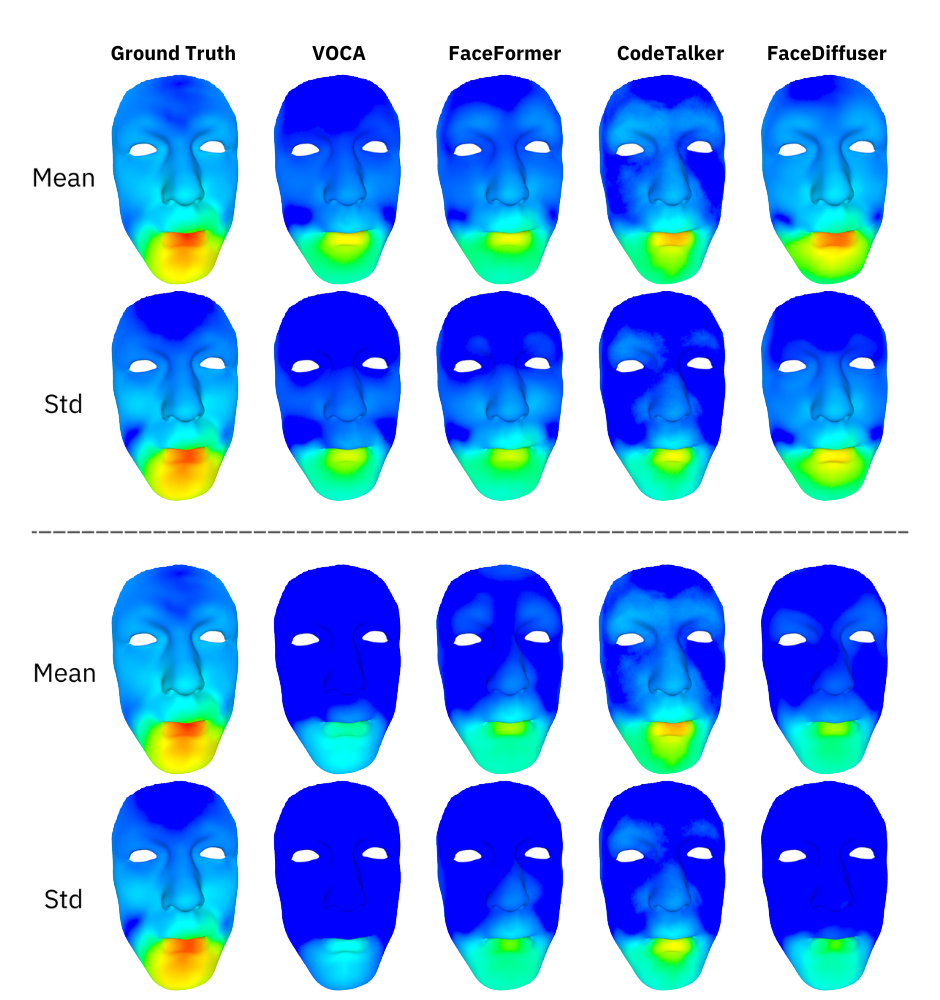}
   \caption{Mean Motion comparison: Animation sequences were sampled from BIWI-Test-B conditioned on different training subjects. One set of two rows separated by the dashed line depicts motion statistics of the same inference. Here we present two sets of inferences generated using the same audio but conditioned on two different training subjects. We notice that conditioning of different subjects produces diversity in generated animations. Whereas, the other methods are much less diverse as seen in the mean motion heatmaps, where dark blue means less observed motion and bright red means more observed motion.}
   \Description{}
   \label{fig:biwi_motion_comparison}
\end{figure}

\begin{figure}
     \begin{subfigure}[b]{0.49\linewidth}
         \includegraphics[width=\linewidth,clip]{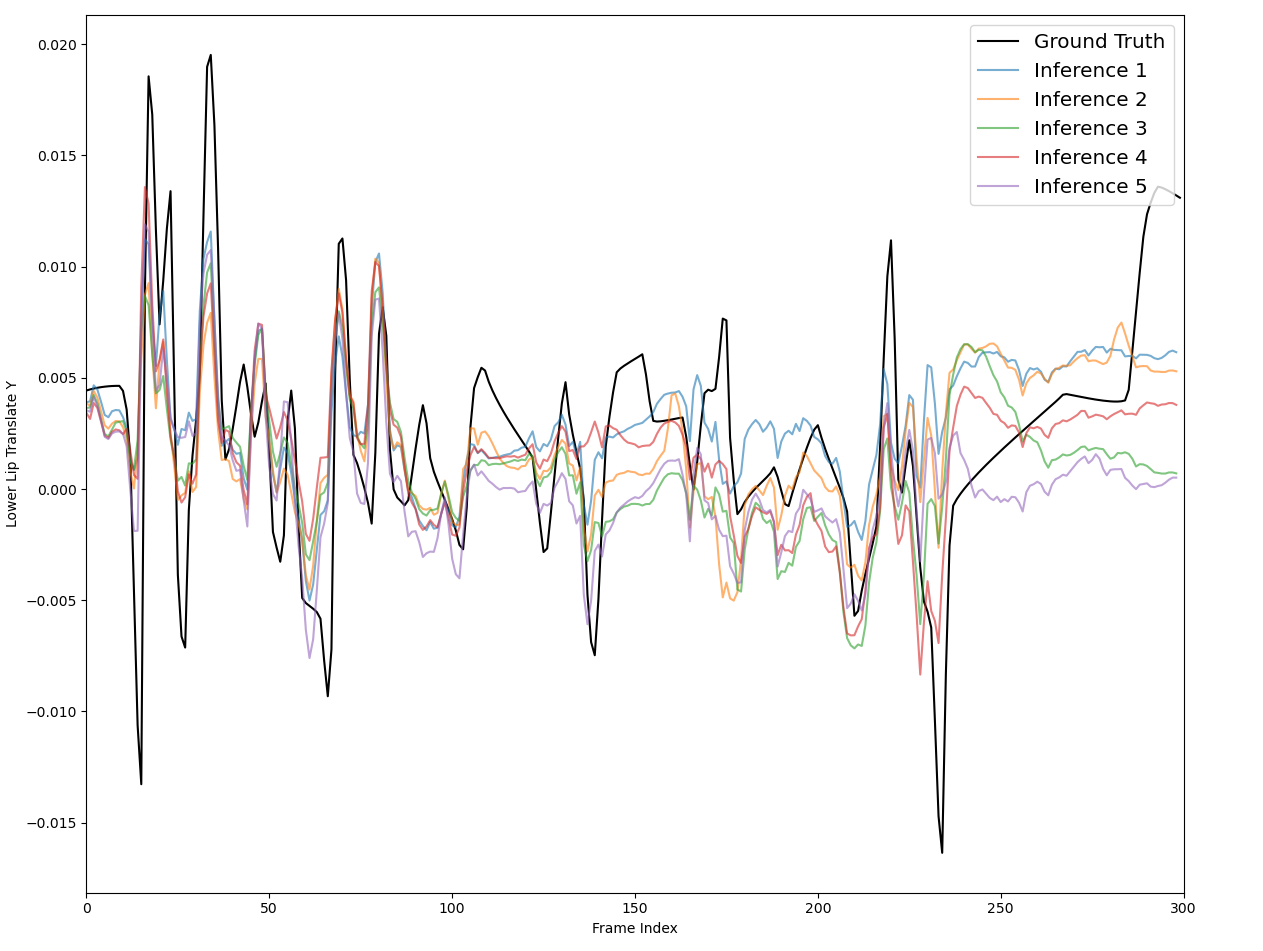}
         \caption[]{Lower lip facial control.} 
         \label{fig:lowerlip}
     \end{subfigure}
     \hfill
     \begin{subfigure}[b]{0.49\linewidth}
         \includegraphics[width=\linewidth,clip]{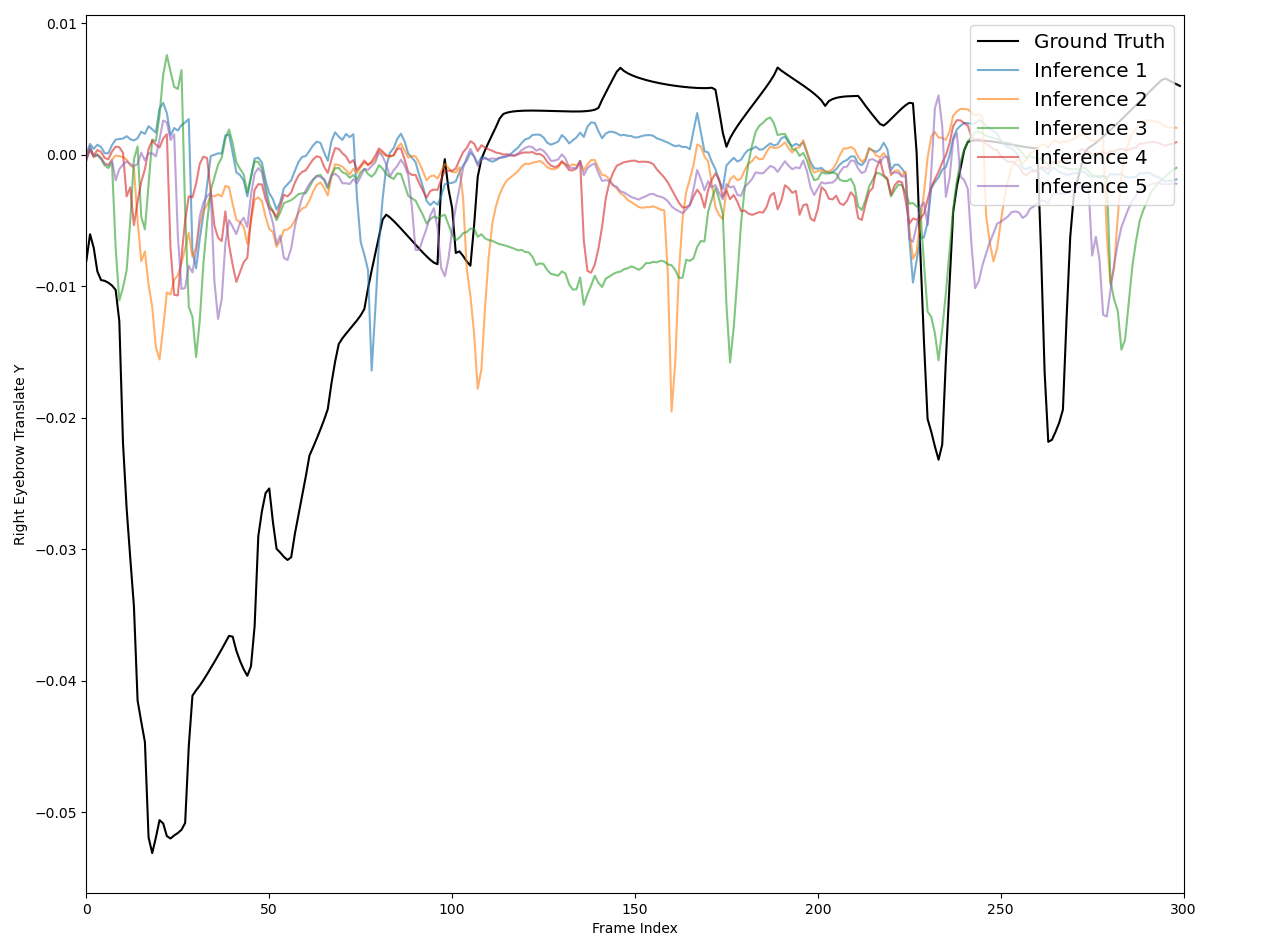}
         \caption[]{Eye brow facial control.} 
         \label{fig:eyebrow}
     \end{subfigure}

     \vskip\baselineskip
     \begin{subfigure}[b]{0.49\linewidth}
         \includegraphics[width=\linewidth,clip]{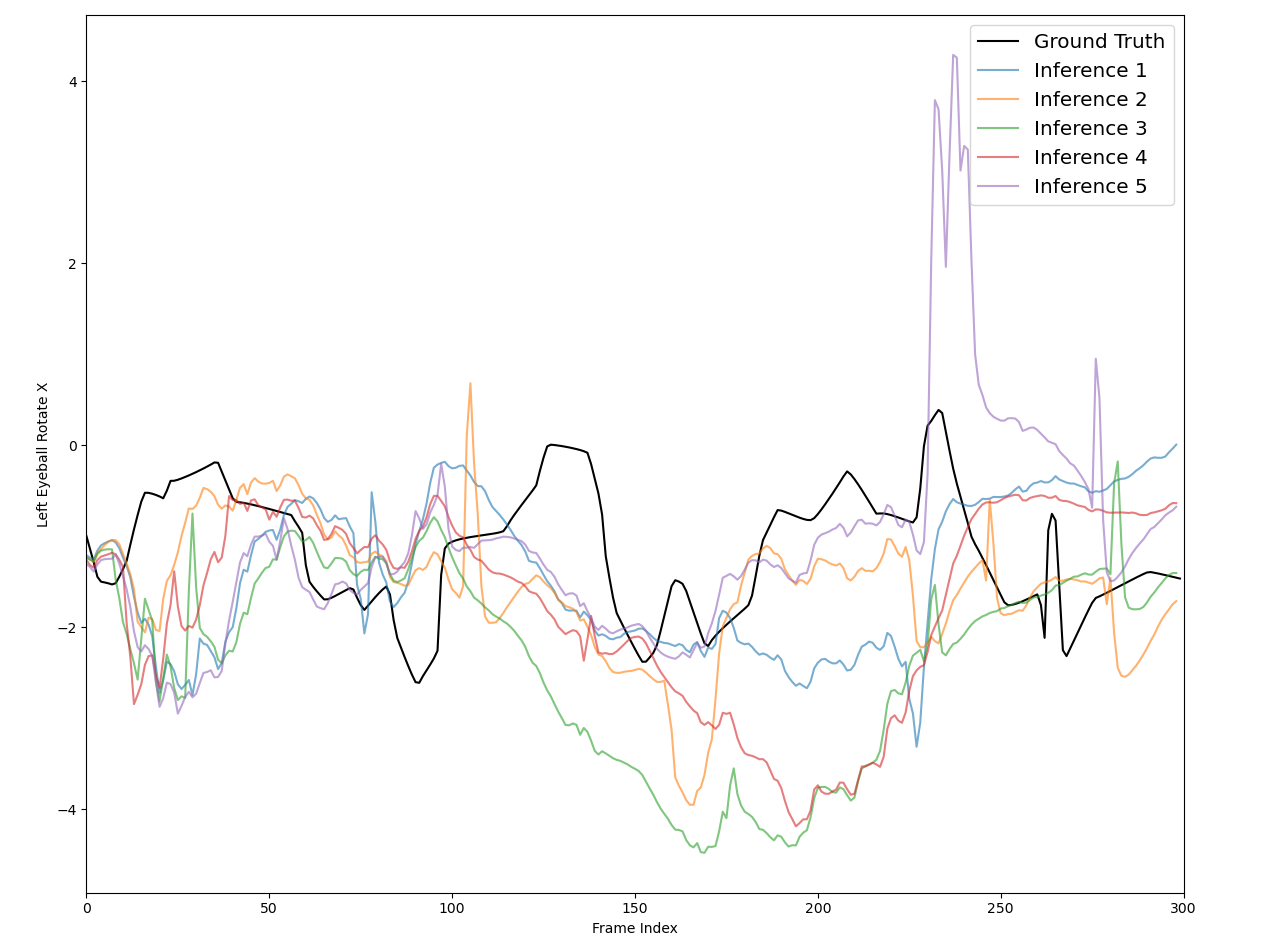}
         \caption[]{Eye ball rotation facial control.} 
         \label{fig:eyeball}
     \end{subfigure}
     \hfill
     \begin{subfigure}[b]{0.49\linewidth}
         \centering
         \includegraphics[width=\linewidth,clip]{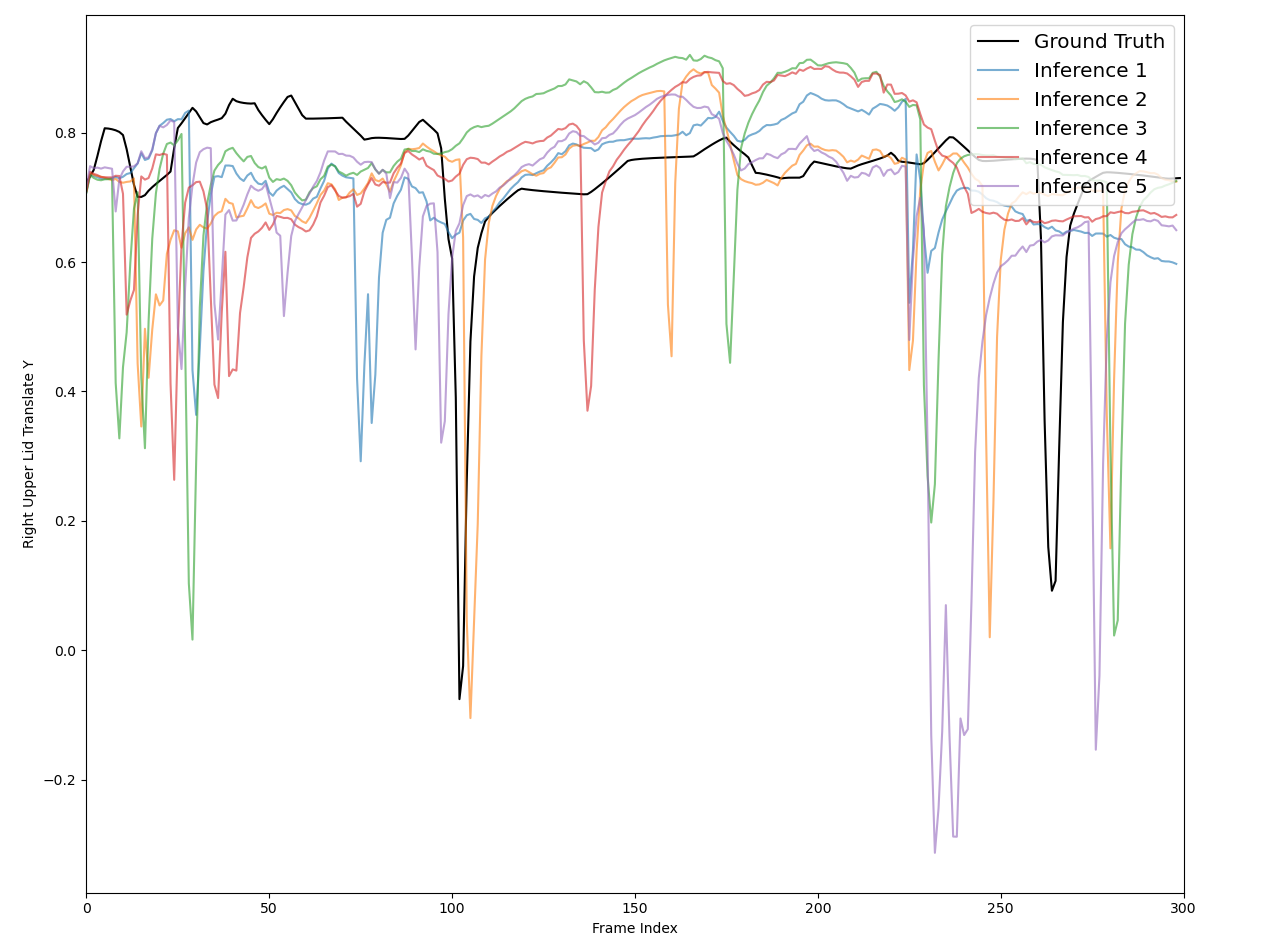}
         \caption[]{Upper lid facial control.} 
         \label{fig:upperlid}
     \end{subfigure}
        \caption{Animation graphs of some facial controls (i.e.- lowerlip, eyebrow, gaze, upperlid) of the UUDaMM dataset. We synthesise animation data multiple times using the same audio and plot the graphs together with the ground truth. The black plots depict the ground truth whereas different coloured plots depict different inferences. It is evident that our approach produces lip control values similar to the ground truth as seen in \cref{fig:lowerlip} while encouraging diversity for the other facial controls as seen in \cref{fig:eyebrow}, \cref{fig:eyeball} and \cref{fig:upperlid}.} 
        \label{fig:DaMM_diversity_graph}
\end{figure}

In order to qualitatively demonstrate the diversity metric presented in \cref{sec:quant_eval}, we sample animation sequences from BIWI-Test-B generated with the same audio but conditioned on different training subjects. Using the generated sequences, we plot the mean and standard deviation of the motion and present it with a heatmap visualisation as shown in \cref{fig:biwi_motion_comparison}. Because there is no subject conditioning for B-FaceDiffuser, we demonstrate the diversity of the generated sequences with animation graphs. We sample our model multiple times using the same audio input and plot the animation graphs of some key facial controls. This can be seen in \cref{fig:DaMM_diversity_graph} where we can observe that upper face controls which are uncorrelated or loosely correlated to speech, do not follow the ground truth whereas the speech correlated lip controls resemble more to the ground truth, especially the peaks in the graph. Furthermore, a high diversity across different results can be observed, especially in the case of eye blinks. For visual judgement, we refer to our supplementary video.

\subsection{Perception Study}
In addition to quantitative and qualitative analyses, we also conducted a series of user studies to evaluate the user perception in terms of realism, lip-sync and appropriateness. We adopt a similar A/B testing strategy for the user study as done in the previous state of the arts-\cite{fan2022faceformer, xing2023codetalker}. We conducted three separate user studies for the compared models trained on three datasets - BIWI, VOCASET and UUDaMM. As visual renders of Multiface resemble VOCASET while BEAT resembles BIWI, these two were dropped for our perception study. For the studies, we used the rendered videos of the generated animation sequences by different methods and presented randomly ordered pairs of videos in a side-by-side manner. The participants were asked to judge three subjective aspects- realism, lip-sync and appropriateness of the rendered animations with respect to the audio. The user studies were hosted on Qualtrics\cite{Qualtrics} and participants were recruited using Prolific\cite{Prolific}, ensuring proper remuneration for their time. Some additional participants were recruited as well who did the studies voluntarily. In total, there were 83 survey responses for the three studies where 31 participated in the study conducted on BIWI, 29 participated in the study on VOCASET and finally, 23 people did the study on UUDaMM. More details about the user studies can be found in the supplementary material. 

\cref{tab:user_study_results} shows the results of the 3 surveys we conducted. Our model clearly outperforms VOCA in all three aspects for both BIWI and VOCASET. However, renders of FaceFormer and CodeTalker were perceived better than ours for BIWI whereas for VOCASET ours perform similarly to FaceFormer while worse than CodeTalker. As reported in \cref{tab:obj_vertex_data}, unlike CodeTalker which produces the same motion conditioned on different training subjects, our model produces diverse animation. This might have affected the user study results. For example, a generated sequence conditioned on a less expressive training subject will have less motion for our model while for CodeTalker, the motion is not affected by the subject condition, see \cref{fig:biwi_motion_comparison}. Resulting in more motion and expressiveness in the rendered videos for CodeTalker that might have affected the perceived subjective aspects when presented side-by-side for this kind of instances. For UUDaMM, there is a clear preference for the results generated by our final model with diffusion than the baseline model without diffusion. For all three datasets, ground truth was perceived as superior, which is expected.

\begin{table}[t]
    \centering
    \begin{tabular}{cccc}
    \toprule
    \multicolumn{4}{c}{Dataset- BIWI}\\
    \midrule
        Competitor & Realism & LipSync & Appropriateness\\
    \midrule
         VOCA & 77.27 \% & 69.32 \% & 79.55\% \\
         FaceFormer & 31.03 \% & 34.48 \% & 37.93 \% \\
         CodeTalker & 44.94 \% & 44.94 \% & 41.57 \% \\
         GT & 38.71 \% & 34.41 \% & 36.56 \% \\
    \midrule
    \multicolumn{4}{c}{Dataset- VOCASET}\\

    \midrule
         VOCA & 76.83 \% & 78.05 \% & 78.05 \% \\
         FaceFormer & 49.38 \% & 46.91 \% & 51.85 \% \\
         CodeTalker & 27.16 \% & 26.40 \% & 27.16 \% \\
         GT & 23.81 \% & 33.33 \% & 27.38 \% \\
         \midrule
         \multicolumn{4}{c}{Dataset- UUDaMM}\\
         \midrule
         w/o Diffusion & 63.77 \% & 66.67 \% & 65.94 \% \\
         GT & 18.48 \% & 27.17 \% & 20.65 \% \\
    \bottomrule
    \end{tabular}
    \caption{User study results. We conduct A/B testing and report the percentage of responses where A (i.e. ours) was preferred over B (i.e. competitor) in terms of realism, lip-sync and appropriateness of the rendered animations.}
    \label{tab:user_study_results}
\end{table}

\subsection{Ablation Study}
To analyse the effects of the different components of our proposed architecture, we experimented with different configurations of it, by either removing or changing different modules. We conducted ablation studies in terms of (i) diffusion mechanism, (ii) audio encoder and (iii) facial decoder. For the ablation study experiments, we only employ BIWI for vertex based dataset, and UUDaMM for blendshape based dataset.

\begin{figure}
\centering
     \begin{subfigure}[b]{0.8\linewidth}
         \includegraphics[width=\linewidth,clip]{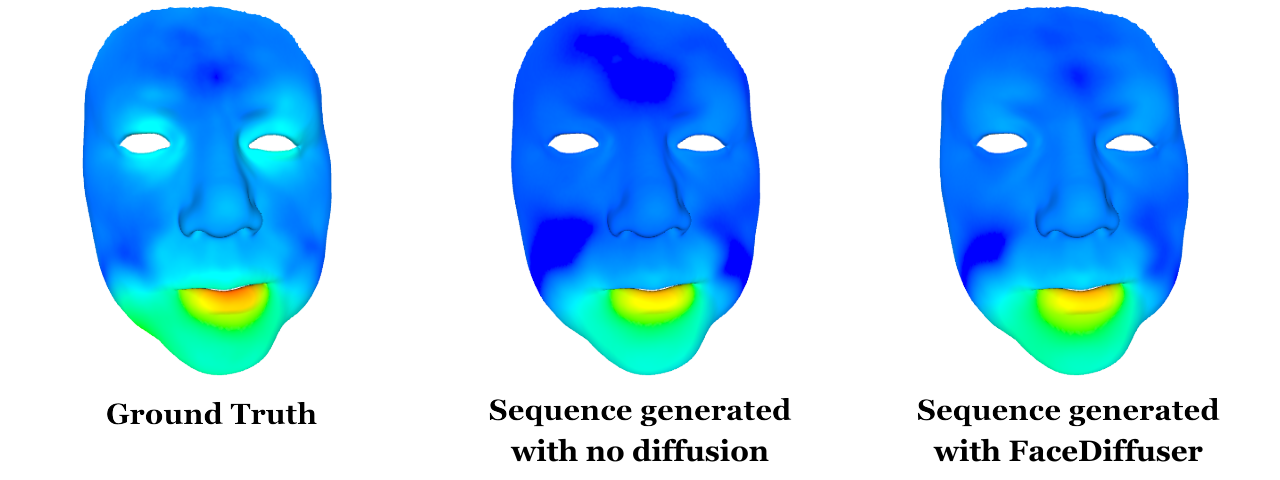}
         \caption[]{Mean motion plot using temporal vertex data of inferences generated without (middle) and with diffusion (right) for BIWI.} 
         \label{fig:BIWI_abl_diff}
     \end{subfigure}
     \begin{subfigure}[b]{0.6\linewidth}
         \includegraphics[width=\linewidth,clip]{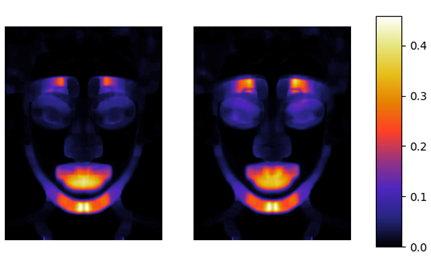}
         \caption[]{Mean motion plot using optical flow of inferences generated without (left) and with (right) diffusion for UUDaMM.} 
         \label{fig:DaMM_abl_diff}
     \end{subfigure}
        \caption{Ablation on the diffusion process. It is evident that the diffusion mechanism encourages more non-verbal cues throughout the face.} 
        \label{fig:abl_diff}
\end{figure}

\textbf{Ablation on Diffusion Process:} To understand the contribution of the diffusion mechanism in our proposed model, we train a similar model without diffusion and compare the results. We experimented with both BIWI and UUDaMM datasets. The first segment of \cref{tab:abl_diff_audio} shows that the model without diffusion achieves slightly worse results in terms of the objective metrics. \cref{fig:abl_diff} shows that models with diffusion produce more motion throughout the face, resembling mean motion observed in ground truth.  

\textbf{Ablation on Audio Encoder:} We use HuBERT as our audio encoder in the proposed model. Following \cite{fan2022faceformer}, a lot of recent works employed Wav2Vec 2.0\cite{baevski2020wav2vec2} as the audio encoder. In order to justify our choice of using HuBERT, we trained our model with Wav2Vec 2.0 as our audio encoder. The second segment in \cref{tab:abl_diff_audio} shows a clear improvement of using HuBERT over Wav2Vec 2.0 to encode audio information for speech-driven downstream tasks of 3D animation synthesis. 

\textbf{Ablation on Decoder:} In order to motivate our choice of the facial decoder in the proposed model, we carried out ablation study experiments by using different sequence modelling method for the decoder. We replaced the proposed GRU decoder with a simpler RNN decoder and more complex transformer decoder and report the objective results in \cref{tab:abl_decoder} for both BIWI and UUDaMM. GRU decoder performs the best out of the tested configurations, resulting in lower error values on both datasets.

\section{Discussion and Limitation}
Our approach performs objectively better than state of the arts but the perceptual evaluation shows that there is still room for improvement in terms of subjective aspects. There is a limitation in terms of available datasets. We argue that the sequences are too short to capture the full range of expressions a person can have in \cite{fanelli20103biwi, wuu2022multiface, cudeiro2019voca}, therefore our model cannot capture long-term context from the data. It would be interesting to see how our model performs against other methods when a larger dataset is used. Furthermore, a dataset comprising diverse captured sequences where for one textual content, subjects perform multiple times, would help to better analyse the diversity capability of non-deterministic models in addition to analysing the non-deterministic nature of non-verbal facial cues uncorrelated or loosely correlated with speech. Only in BIWI, this is available but only twice in terms of binary emotion conditions. Incorporating emotion information can be a future direction to explore, including categorical or continuous emotion models. Using BEAT dataset or creating synthetic datasets using vision-based 4D reconstruction and emotion recognition models such as EMOCA\cite{EMOCA:CVPR:2021} similar to \cite{emote}, can be a potential future direction to this end. Since our results have limitations in terms of perceived realism, employing a diffusion autoencoder similar to \cite{du2023dae} can prove to be a potential future direction to achieve better quality 3D facial animation synthesis in terms of the subjective aspects. Moreover, it would be interesting to see how models proposed for vertex based datasets perform when trained on lower dimensional blendshape datasets. Furthermore, due to the iterative sampling process of the diffusion mechanism, our model's inference time is long and subject to the diffusion timesteps used during training. Therefore, our approach is not suitable for real-time applications.

\begin{table}[t]
    \centering
    \resizebox{\linewidth}{!}{
    \begin{tabular}{ccccc}
    \toprule
    \multicolumn{5}{c}{Ablation on Diffusion Process}\\
    \midrule
        Model & MVE $\downarrow$ & LVE $\downarrow$ & FDD $\downarrow$ & Training \\
         & x$10^{-3}$mm & x$10^{-4}$mm  & x$10^{-5}$mm & Time (m)\\
    \midrule
        w/o Diffusion & 6.8833 &  4.5870 & 4.6690  & $\approx 67$\\
        FaceDiffuser & 6.8088 & 4.2985 & 3.9100  &  $\approx 67$ \\
    \midrule
    \multicolumn{5}{c}{Ablation on Audio Encoder}\\
    \midrule
    Model & MVE $\downarrow$ & LVE $\downarrow$ & FDD $\downarrow$ & Training \\
         & x$10^{-3}$mm & x$10^{-4}$mm  & x$10^{-5}$mm & Time (m)\\
    \midrule
    Wav2Vec2 & 7.4593 & 5.1590 & 4.1950 & $\approx 67$  \\
    HuBERT & \textbf{6.8088} & \textbf{4.2985} & \textbf{3.9100} & $\approx 67$  \\
    \bottomrule
    \end{tabular}
    }
    \caption{Objective metrics computed over the test results of BIWI dataset for ablation experiments of the diffusion process and of different audio encoder.}
    \label{tab:abl_diff_audio}
\end{table}

\begin{table}[t]
    \centering
    \begin{tabular}{ccccc}
    \toprule
        \multicolumn{5}{c}{Dataset- BIWI}\\
    \midrule
        Decoder & MVE $\downarrow$ & LVE $\downarrow$  & FDD $\downarrow$  & Training \\
        Type &  x$10^{-3}$ &  x$10^{-4}$  & x$10^{-5}$   & Time \\
         &   mm &  mm & mm  & (h)\\
    \midrule
        GRU & 6.8088 & 4.2985 & 3.9100  &  $\approx 1.12$ \\
        RNN & 7.0833 &  4.7870 & 4.0690  & $\approx 1.12$\\
        Transformer(TF)& 9.9767 & 10.1300 & 4.8587  & $\approx 1.34 $\\
        Transformer(AR) & 7.0213 & 4.6941 & 4.3272  & $\approx 5.00$\\
    \midrule
    \multicolumn{5}{c}{Dataset- UUDaMM}\\
    \midrule
        Decoder & MBE $\downarrow$ & LBE $\downarrow$  & FDD $\downarrow$  & Training \\
        Type &  &  &  & Time (h) \\
    \midrule
        GRU & \textbf{3.6791} & \textbf{3.5812} & 1.8862  & $\approx 4.5$ \\
        RNN & 3.8654 & 3.9196 & \textbf{1.8426} & $ \approx 2.92 $\\
        Transformer(AR) & 3.6881 & 3.7105 & 1.8533 & $ \approx 4.58$\\
    \bottomrule
    \end{tabular}
    \caption{Objective metrics computed over the BIWI test results for ablation experiments of facial decoder. Here, (TF) and (AR) depict teacher-forcing scheme and autoregressive scheme respectively for Transformer decoders.}
    \label{tab:abl_decoder}
\vspace{-5mm}
\end{table}

\section{Conclusion}
We integrated the diffusion mechanism into a generative deep neural network trained to generate 3D facial animations conditioned on speech. The proposed approach is generalisable to both high dimensional temporal 3D vertex data as well as low dimensional blendshape data with minimal changes. The quantitative analysis shows that our approach performs better than the state of the arts. We showed that our model is able to produce higher diversity of motions between different style conditions than the competitors. Our approach also produces diverse animation sequences for rigged characters that can be observed in animation graphs of multiple generations conditioned on the same audio. Extensive ablation studies support our network architecture design choices, showing the benefits of different proposed components of the neural network architecture.

\noindent\textbf{Ethical Consideration} Face data can be used for generating content that can jeopardise privacy. We must act responsibly by considering the aspects pertaining to privacy and ethics.

\noindent\textbf{Acknowledgement} We thank the authors of MDM, VOCA, FaceFormer, CodeTalker, BIWI, BEAT and Multiface for making their codes and datasets available. Many thanks to Camille Gruter, Teodor Nikolov and Mihail Tsakov for the hard work acquiring the UUDaMM dataset.

\bibliographystyle{ACM-Reference-Format}
\bibliography{bibtex}

\clearpage

\appendix

\section{Supplementary Material}

\subsection{User studies}
In total, 83 survey responses were collected spread among the 3 experiments as follows- $31$ for the BIWI survey, $29$ for the VOCASET survey, and $23$ for the UUDaMM survey. An additional $3$ responses were discarded due to the participants failing the attention checks. The surveys were distributed online to different groups that voluntarily took part in the study without being remunerated, managing to collect $38$ responses out of the total. The additional $45$ responses were collected through the Prolific platform, which facilitates response collection by allowing participants to take part in surveys and get remuneration. We fairly compensated the participants by awarding them the equivalent of $9 \pounds / h $. Most of the participants are adults or young adults, with the following age distribution: $65.85 \%$ in the 18-25 age group, $25.61$ in the 26-35 age group, and $8.54 \%$ in the 36-45 age group. Looking at the gender distribution, $36.59 \%$ of the participants identify as female, $54.88 \%$ as male, and $8.53 \%$ as non-binary/other.

In regards to the user familiarity with the subject, we computed the average reported familiarity of the 3 areas that were questioned and we obtained $2.71$ average familiarity with virtual humans, $3.52$ average familiarity with 3D animated movies, and $4.06$ average familiarity with video games on a 5 point likert scale. 

\begin{figure}[t]
\centering
     \begin{subfigure}[b]{0.95\linewidth}
         \includegraphics[width=\linewidth,clip]{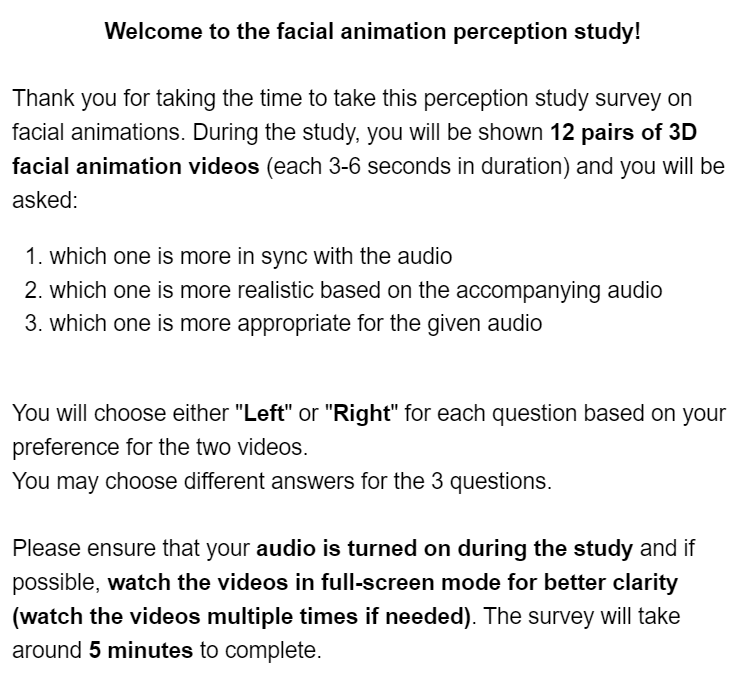}
         \caption[]{User study instruction.} 
         \label{fig:user_study1}
     \end{subfigure}
     \begin{subfigure}[b]{0.95\linewidth}
         \includegraphics[width=\linewidth,clip]{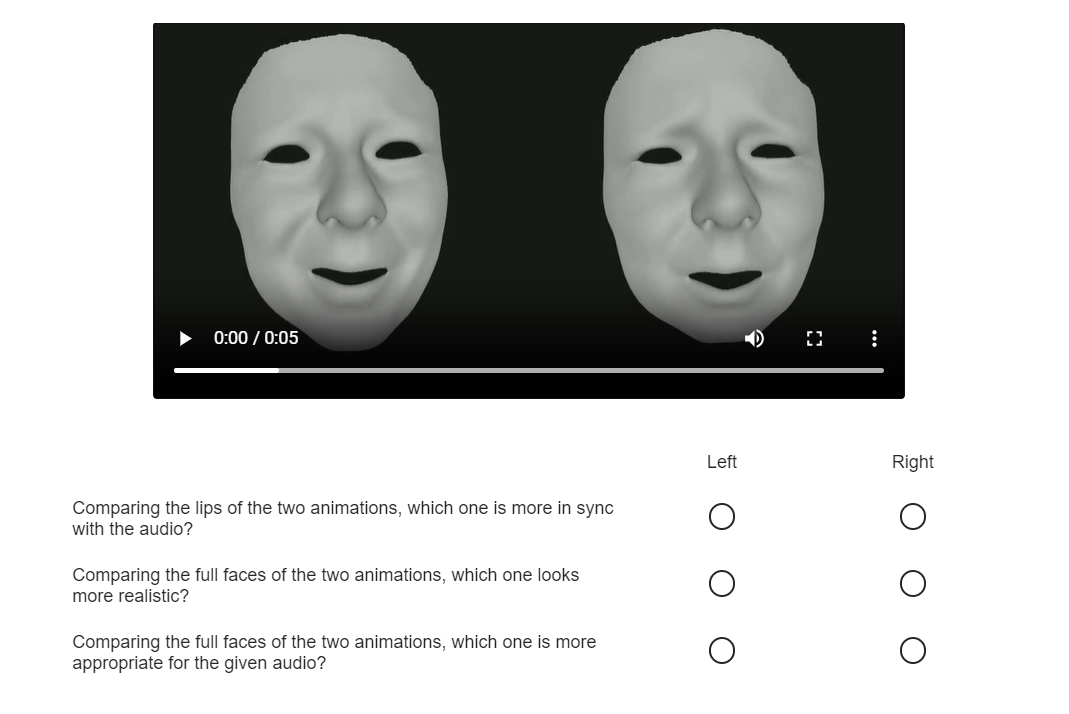}
         \caption[]{User study UI.} 
         \label{fig:user_study2}
     \end{subfigure}
        \caption{Example screenshots of the user studies.} 
        \label{fig:user_study_supp}
\end{figure}

The first 2 surveys present users with 12 pair-wise comparisons. To avoid the impact of random selection by users, we randomly switch the side on which we show our model with our competitors. For each comparison, the user is asked 3 questions. For the first 2 questions we follow previous works \cite{facexhubert2023, fan2022faceformer, xing2023codetalker} and ask the users about lip-sync quality and perceived realism of the animations. Additionally, we add an extra question asking about the animation appropriateness. Survey instruction and an example of the user interface can be seen in \cref{fig:user_study1} and \cref{fig:user_study2} respectively.

The questions the users were presented with are as follows:
\begin{enumerate}
    \item Comparing the lips of the two animations, which one is more in sync with the audio?
    \item Comparing the full faces of the two animations, which one looks more realistic?
    \item Comparing the full faces of the two animations, which one is more appropriate for the given audio?
\end{enumerate}

\subsection{Datasets}
\begin{table*}[t]
    \centering
    \begin{tabular}{lccc}
    \toprule
        Dataset & BIWI & VOCASET & Multiface \\
    \midrule
        Training Set 
        & \specialcell[l]{
        \tabitem 6 subjects \\ \tabitem 32 sequences per subject
        \\ \tabitem Total = 192 sequences}
        & \specialcell[l]{
        \tabitem 8 subjects \\ \tabitem 40 sequences per subject \\ \tabitem Total = 320 sequences}
        & \specialcell[l]{
        \tabitem 9 subjects \\ \tabitem 40 sequences per subject \\ \tabitem Total = 360 sequences} \\
        \hline

        Validation Set 
        & \specialcell[l]{
        \tabitem 6 seen subjects \\ \tabitem 4 sequences per subject
        \\ \tabitem Total = 24 sequences}
        & \specialcell[l]{
        \tabitem 2 unseen subjects \\ \tabitem 20 sequences per subject \\ \tabitem Total = 40 sequences}
        & \specialcell[l]{
        \tabitem 9 seen subjects \\ \tabitem 5 sequences per subject \\ \tabitem Total = 45 sequences} \\
        \hline
        Test Set A 
        & \specialcell[l]{
        \tabitem 6 seen subjects \\ \tabitem 4 sequences per subject
        \\ \tabitem Total = 24 sequences}
        & -
        & \specialcell[l]{
        \tabitem 9 seen subjects \\ \tabitem 5 sequences per subject \\ \tabitem Total = 45 sequences} \\
        \hline
        Test Set B
        & \specialcell[l]{
        \tabitem 8 unseen subjects \\ \tabitem 4 sequences per subject \\ \tabitem 6 conditions per sequence
        \\ \tabitem Total = 192 sequences}
        & \specialcell[l]{
        \tabitem 2 unseen subjects \\ \tabitem 20 sequences per subject \\ \tabitem 8 conditions per sequence \\ \tabitem Total = 320 sequences}
        & \specialcell[l]{
        \tabitem 4 unseen subjects \\ \tabitem 5 sequences per subject \\ \tabitem 9 conditions per sequence\\ \tabitem Total = 180 sequences} \\
    \bottomrule
    \end{tabular}
    \caption{Summary of the dataset split for the 3 vertex based datasets used in our work.}
    \label{tab:dataset_summary}
\end{table*}
\begin{table*}[t]
    \centering
    \begin{tabular}{lcc}
    \toprule
        Dataset & UUDaMM & BEAT \\
    \midrule
        Training Set 
        & \specialcell[l]{
        \tabitem 2 subjects \\ \tabitem 2029 10-second sequences per subject \\ 
        \tabitem Total: 4058 sequences
        }
        & \specialcell[l]{
        \tabitem 4 subjects \\ \tabitem 2175 10-second sequences} \\
        \hline

        Validation Set 
        & \specialcell[l]{
        \tabitem 2 subjects \\ \tabitem 254 10-second sequences per subject \\ 
        \tabitem Total: 508 sequences
        }
        & \specialcell[l]{
        \tabitem 4 subjects \\ \tabitem 274 10-second sequences} \\
        \hline
        Test Set
        & \specialcell[l]{
        \tabitem 2 subjects \\ \tabitem 254 10-second sequences per subject \\ 
        \tabitem Total: 508 sequences
        }
        & \specialcell[l]{
        \tabitem 4 subjects \\ \tabitem 275 10-second sequences}
        \\
    \bottomrule
    \end{tabular}
    \caption{Summary of the dataset split for the 2 blendshape based datasets used in our work.}
    \label{tab:dataset_summary_rig}
\end{table*}
Here, we describe more in detail the datasets that were used in our work. A summary of the datasets can be found in \cref{tab:dataset_summary} and \cref{tab:dataset_summary_rig}. 
\subsubsection{BIWI\cite{fanelli20103biwi}}
The dataset is comprised of $14$ x $40$ x $2$ sequences of paired audio and animation. For the creation of these sequences, $14$ subjects were asked to read and emote $40$ different sentences, each sentence being read $2$ times, one time with a neutral expression and one time with a more emotional one. Each sequence is on average approximately $5$ seconds long and is captured at $25 fps$. The face meshes included are very high-definition comprising $23370$ 3D vertices, despite only the front of the head being depicted. Based on previous work, we use the same data splits as in \cite{fan2022faceformer, xing2023codetalker} and only use the emotional subset of sequences. During training, only 6 subjects (3 female and 3 male) are used, along with 32 spoken sentences per subject. This amounts to a total of $192$ sequences and represents the BIWI-Train dataset. From the remaining 8 sentences of these subjects, $4$ are used for validations ($24$ in total), and $4$ for testing ($24$ in total). We refer to this test set as BIWI-Test-A and will be used to compute objective metrics over our results. With the remaining, 8 subjects and their last 4 sentences, BIWI-Test-B is formed. This dataset represents the primary one we use during the model experiment phase. Furthermore, the hyperparameter tuning along with other experiments and the ablation study are mainly performed on this dataset. 

\subsubsection{VOCASET\cite{cudeiro2019voca}}.
The dataset is comprised of 480 sequences of audio-visual pairs which amount to a total length of just 29 minutes. The sequences are recorded at $60fps$ and the facial scans are translated onto the FLAME head topology \cite{li2017flame} which is comprised of $5023$ 3D vertices. Unlike, the BIWI dataset, the mesh includes the whole head and neck of the person, including eyeballs and eyelids. Even though this would technically allow for more expressivity and possibly even eye blinks to be captured, these are scarcely available in the dataset. In terms of data split, we utilise the one proposed by the authors of the dataset and later used by other methods as well \cite{fan2022faceformer, xing2023codetalker}, 8 of the 12 subjects along with all of their sentences are used for training (320 sequences), 2 subjects with 20 sentences per subject are used for validation (40 sequences), and similarly the last 2 subjects with the last 20 sentences are used for testing (40 sequences). This dataset is used mostly for validating our approach in terms of generalizability and for conducting the user study.

\subsubsection{Multiface\cite{wuu2022multiface}}
The third vertex based dataset in our research, namely the Multiface dataset, publicly released by \textit{Wuu et al.}\cite{wuu2022multiface}. To the best of our knowledge, we are the first ones to use it for the task of facial animation synthesis besides its creators. In comparison to the other datasets, the face meshes included here are more complex in terms of features, containing attributes such as hair, eyelids and facial hair. Moreover, the dataset includes full 3D head scans, including the back of the head as well as the neck.

The publicly released version of the dataset contains a total of 13 subjects out of 250 subjects that was used for training Meshtalk \cite{richard2021meshtalk}. Even though the full dataset is much larger, comprising a total of $250$ subjects, it is not available to the public. For each subject, there are a total of 50 spoken sentences available. Since the sequences are split by subject and sentence, it allows us to have a similar training technique, including the one-hot style embedding. The authors of the dataset share that the sentences were chosen in such a way that they are phonetically balanced ensuring a good generalisation across the possible phonemes. The animation is sequence of 3D face meshes available at $30$ frames per second. Each frame is represented by the full 3D face of the actor, with a total of $6172$ 3D vertices, including eyelids, neck, as well as different hairstyles for the subjects. Since there is no previous work to follow for this subset of the dataset, we propose our own data split (similar BIWI in terms of number of sequences) and use 9 of the subjects with the first 40 sentences for training (360 sequences) and call this Multiface-Train, the following 5 sentences of the same 9 subjects are used for validation (45 sequences), and the last 5 sentences make up Multiface-Test-A and are used for testing (45 sequences). With the other 4 subjects and their last 5 sentences that were unseen during training, we form Multiface-Test-B.

\subsubsection{Utrecht University Dyadic Multimodal Motion Capture Dataset (UUDaMM)}
Our in-house UUDaMM dataset consists of synchronously captured dyadic conversations between 2 actors in terms of four modalities- gesture, face, audio and text. The dataset contains 9 hours and 41 minutes of recorded conversations between two speakers, of which 6 hours and 53 minutes represent active speech sequences, in which at least one of the two actors is speaking. For our work in hand, we only discuss on the facial data in this document, but we consider that the way the dataset was collected, is also helpful for the task of facial animation via multi-modal learning approaches as well as for modelling dyadic interaction.

The facial performance capture, done with Dynamixyz\cite{Dynamixyz} system at 120 FPS, is solved and retargeted to a virtual 3D character in a Maya scene. We use a publicly available character\cite{RayCharacterMaya} to ensure ease of use among the research community. The character comes with artist generated blendshapes as well as intuitive facial controls that drive the blendshapes. After the facial performance is solved and retargeted, we use a python script to extract the temporal facial control values to form the training dataset. A similar python script can be used to import the inferred data into the Maya scene as well. 

\subsubsection{BEAT\cite{liu2021beat}}
The second blendshape based dataset we use is the BEAT dataset \cite{liu2021beat}. Just like UUDaMM, the dataset also contains body motion data along with facial capture, the difference being that the facial animations are encoded as sequences of blendshape weights instead of facial controls.

iPhone 12 Pro is used to capture the facial data of the actors, encoded as 52 blendshape weights defined in Apple's ARKit. The frame rate of the facial capture is 60 FPS. The full dataset includes 30 participants, of which half are female and half male. Each participant was asked to read 118 predefined texts, each resulting in a one-minute recording, in various emotional ways in order to cover multiple variations of emotional speech. The 8 emotions captured during the collection are neutral, happiness, anger, sadness, contempt, surprise, fear, and disgust. An additional 12 recordings of 10 minutes each were captured for each participant in which they perform free-form conversations with an off-screen director. Furthermore, the dataset contains sequences spoken in 4 different languages: English, Chinese, Japanese, and Spanish, also including native and non-native English speakers. The total size of the recordings amounts to about 76 hours. For our training, we use a 16 hour subset of the dataset comprising native English speaking sequences of 4 subjects. All these features make the dataset to be the most diverse out of the ones we consider, the only downfall being that "true" dyadic conversations are not recorded. 
In terms of pre-processing, we do not apply any transformations to the values of the features themselves, and merely split the sequences so that the data follows a similar format to that present in the other datasets. After splitting the provided sequences into 10-second segments, we obtain a total of $11398$ that are used for training. In terms of data split, we employ an 80-10-10, training-validation-test split.

\begin{table}[t]
    \centering
    \begin{tabular}{ccccc}
    \toprule
        Diffusion & MVE $\downarrow$ & LVE $\downarrow$ & FDD $\downarrow$ & Diversity $\uparrow$ \\
        Steps & x$10^{-3}$mm & x$10^{-4}$mm  & x$10^{-5}$mm &  \\
    \midrule
        100 & 7.2391 & 4.7703 & 5.0705 & 0.8236 \\
        250 & 6.9618 & 4.5515 & 4.2407 & 0.8331 \\
        500 & \textbf{6.8507} & \textbf{4.4364} & 4.3212 & 0.8446 \\
        750 & 7.1290 & 4.6428 & \textbf{4.1268} & \textbf{0.8725} \\
        1000 & 7.0387 & 4.6897& 4.5889 & 0.8617\\
    \bottomrule
    \end{tabular}
    \caption{Evaluation metrics computed over the test results of different numbers for the diffusion timesteps}
    \label{tab:diff_steps}
\end{table}

\subsection{Additional Ablation on Diffusion Steps}
We experimented with different diffusion step numbers as can be seen in \cref{tab:diff_steps}. Analysing the visual results we observe that the number of diffusion steps does not influence the model's capacity of producing acceptable animations, all of the configurations producing correct lip-sync animations. The differences are mostly noticed when it comes to the general expressivity of the animations. The results obtained with just $100$ diffusion steps are generally the worst both objectively and subjectively, while we do not see a lot of visual differences between the results obtained with the other tested values. Both $500$ and $750$ diffusion time steps yield acceptable results, however increasing the value beyond that worsens the results.

\begin{figure}[t]
  \centering
   \includegraphics[width=0.95\linewidth]{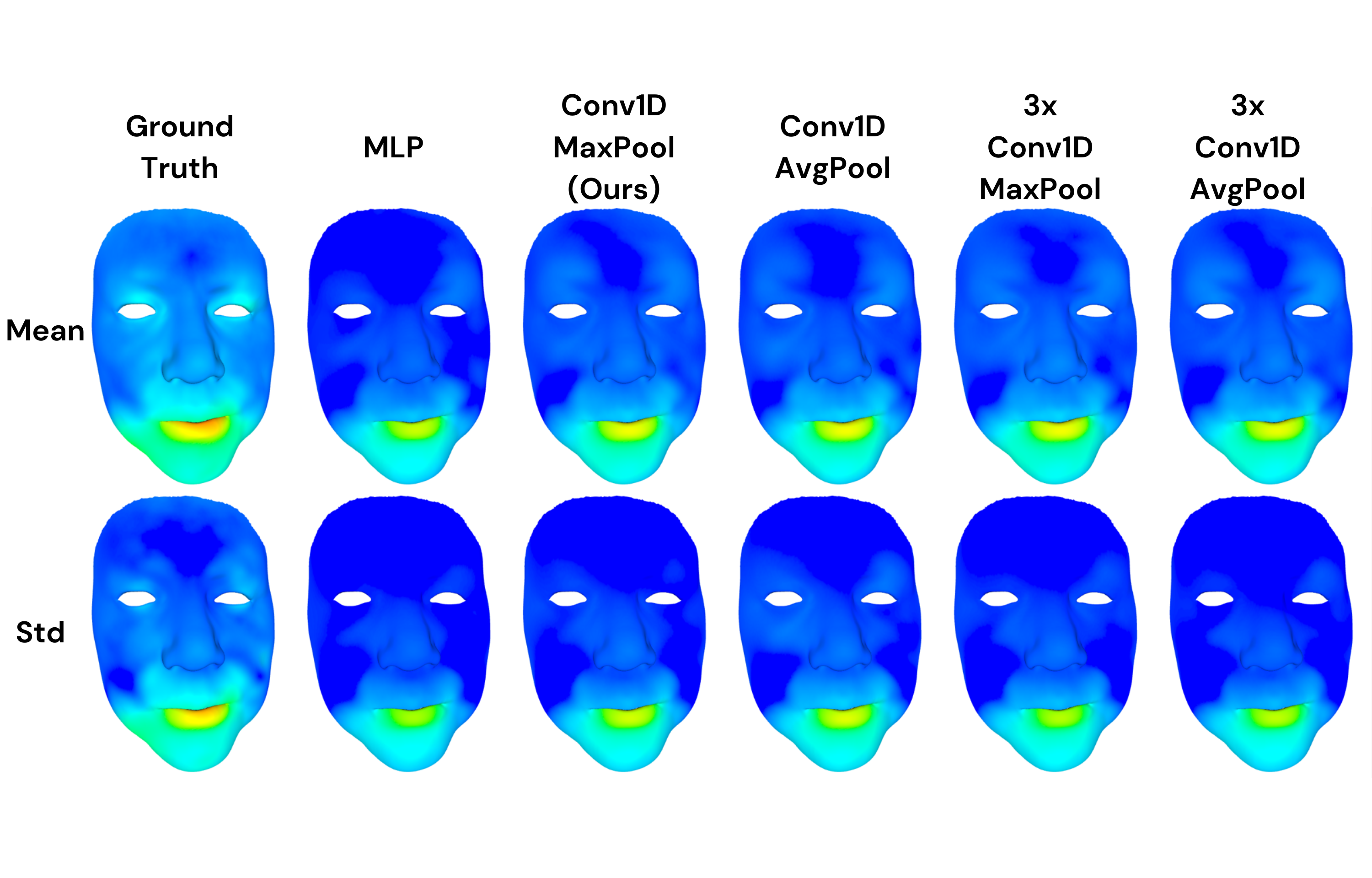}
   \caption{Ablation on the noise encoder.}
   \Description{}
   \label{fig:ablation_noise_encoder}
\end{figure}

\begin{table}[t]
    \centering
    \resizebox{\linewidth}{!}{
    \begin{tabular}{cccc}
    \toprule
        Noise & MVE $\downarrow$ & LVE $\downarrow$ & FDD $\downarrow$ \\
        Encoder & x $10^{-3}$mm & x $10^{-4}$mm  & x $10^{-5}$mm \\
    \midrule
        MLP & 7.1728 & 4.9453 & 3.7748 \\
        Conv1D + MaxPool (Ours) &  \textbf{6.8088} & \textbf{4.2985} & 3.9100 \\
        Conv1D + AvgPool &  6.8735 & 4.5766 & 4.2383 \\
        3 x (Conv1D + MaxPool) & 6.9217 & 4.5241 & \textbf{3.6020} \\
        3 x (Conv1D + AvgPool) & 6.8415 & 4.4130 & 3.9621 \\
    \bottomrule
    \end{tabular}
     }
    \caption{Results of different types of encoder for the noise}
    \label{tab:abl_noise}
\end{table}

\subsection{Additional Ablation on Noise Encoder}
For V-FaceDiffuser, due to the large number of 3D vertices in the data, we introduced a noise encoder in order to reduce the high dimensionality of the noise input to a low dimensional latent representation. We experimented with different configurations for the noise encoder as follows-

\begin{itemize}
    \item \textbf{MLP.} A series of 2 fully connected layers.
    \item \textbf{Conv1D + MaxPool (Ours)} One fully connected layer followed by a single-dimensional convolution and a max pooling layer.
    \item \textbf{Conv1D + AvgPool } One fully connected layer followed by a single-dimensional convolution and an average pooling layer.
    \item \textbf{3 x Conv1D + MaxPool} Same as the second item but three times.
    \item \textbf{3 x Conv1D + AvgPool } Same as the third item but three times.
\end{itemize}

By analysing the results in both \cref{tab:abl_noise} and the heatmaps in \cref{fig:ablation_noise_encoder}, we can see a trend in the expressivity of the animations, viewed as a higher activation of the face, can be seen as the noise encoder becomes more complex. Some more complex configurations (i.e. row 4 and 5 in \cref{tab:abl_noise}) were also tried, showing better results in terms of the mean motion of the face. However, these come at the detriment of the error metrics compared to ground truth data. Out of the experimented configurations, our proposed choice performed the best. We can also notice that the Conv1D approaches are performing better than the MLP, both by looking at the results in \cref{tab:abl_noise} and the visual representation in \cref{fig:ablation_noise_encoder}. Furthermore, the method using max pooling performs slightly better than the average pooling one.

\subsection{Additional Use Cases}
\subsubsection{Animation Editing}
\begin{figure}[t]
\centering
     \begin{subfigure}[b]{0.95\linewidth}
         \includegraphics[width=\linewidth,clip]{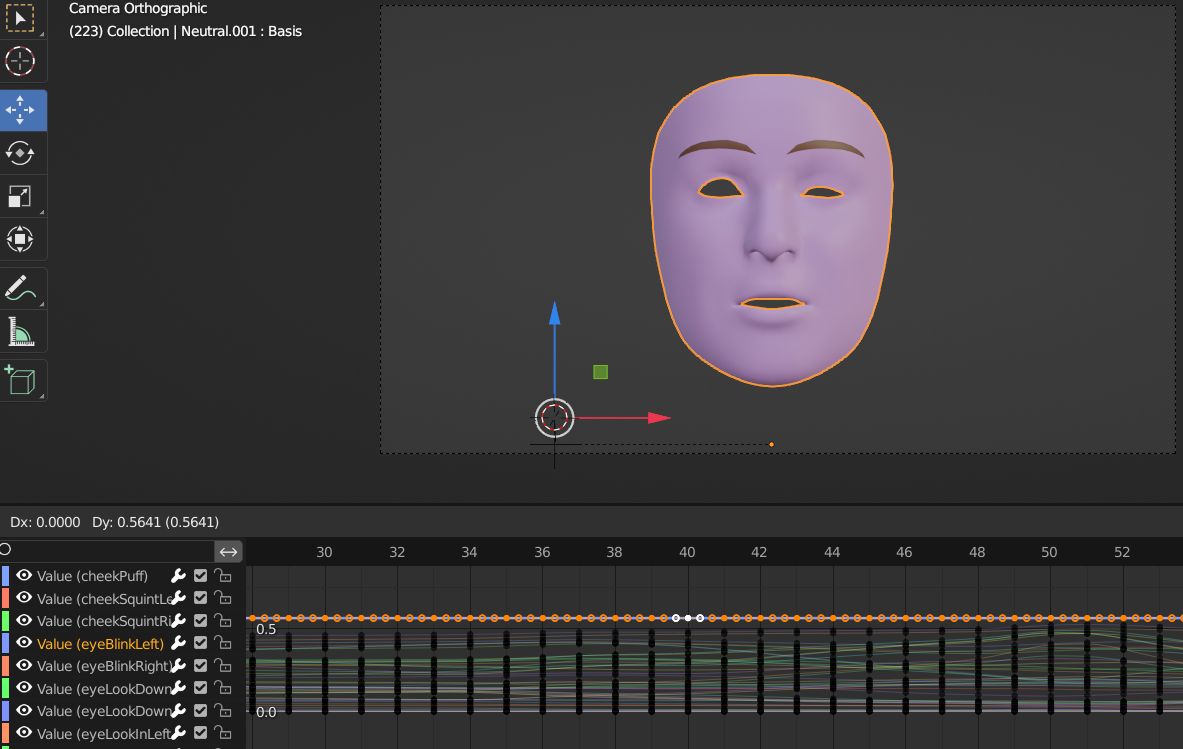}
         \caption[]{An example of how the generated animation can be edited by an animator after automatically generated by our model.} 
         \label{fig:blender_edit}
     \end{subfigure}
     \begin{subfigure}[b]{0.95\linewidth}
         \includegraphics[width=\linewidth,clip]{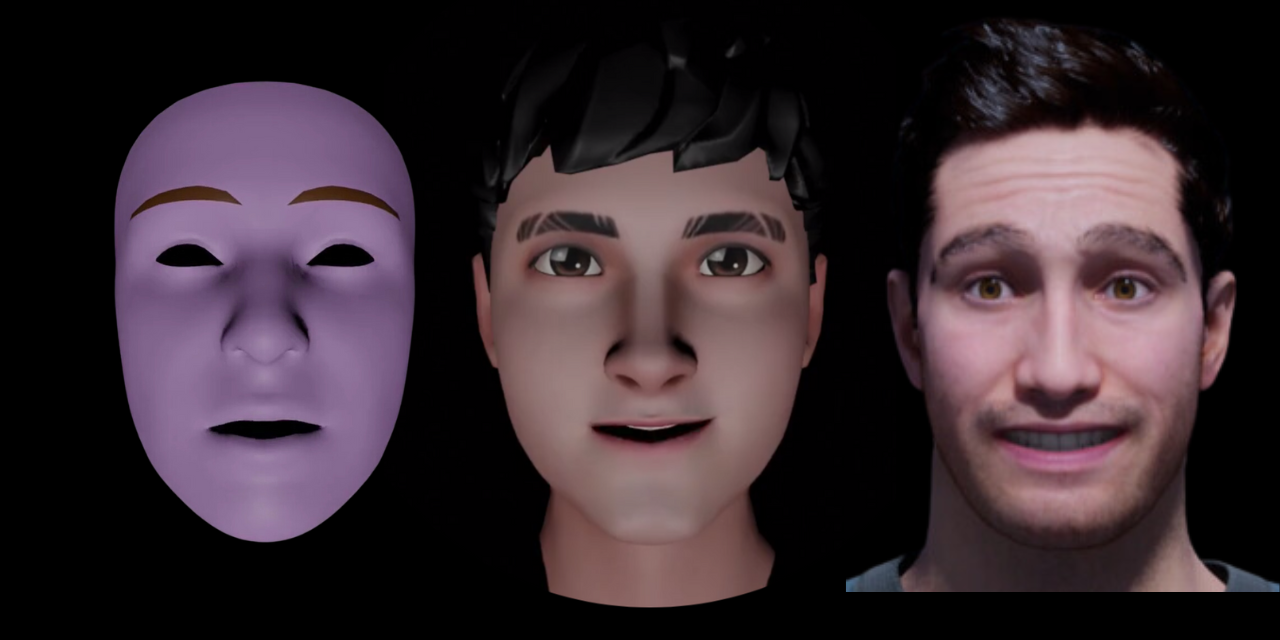}
         \caption[]{By using the BEAT dataset, FaceDiffuser is able to generate ARKit blendshape animations that can be used to animate a large variety of 3D characters that are ARKIT Blendshape enabled such as Ready Player Me avatars and Epic Games' MetaHumans.} 
         \label{fig:beat_application}
     \end{subfigure}
        \caption{Use cases of our approach in existing animation production workflows.} 
        \label{fig:metahuman_anim}
\end{figure}

By using a generalised animation encoding such as the Apple ARKit blendshape expression space, we are able to easily edit the resulting animations to better emulate our desired results. This can be done by simply updating the animation curves as can be seen in \cref{fig:blender_edit} and in the accompanying supplementary video. A simple change like moving one of the curves upwards would change the entire sequence expression and could be used to generate various expressions. For example, an animator might choose to generate more obvious mouth movements by applying a filter over the \textit{mouthOpen} blendshape, which would make mouth opens and closures more extreme.
Considering that accurate lip movements are automatically generated by a data-driven model like ours, the animators then have freedom of customising those animations to better fit their desires.

\subsubsection{Animation Transfer}
Another identifiable use case of generating ARKit blendhshape animations (with the model trained on BEAT dataset) is the transferability of such animations between different characters having the same blendshapes or semantically the same/similar blendshapes with different names. No additional retargeting steps are required if the blendshape names are the same as ARKit ones while a trivial script that maps the blendshape names between source data and target data would solve the retargeting and therefore the transfer is easy even for novice animators. This opens a wide-range of opportunities as there is a vast array of different faces that can be animated by using our model. However, due to the limited capability in terms of expressiveness in ARKit blendshapes, applying them to high fidelity photorealistic human characters may cause the effect known as the uncanny valley effect.

\end{document}